\documentclass[sigconf]{acmart}
\usepackage{amsmath}
\usepackage{amsfonts}
\usepackage[linesnumbered,ruled,vlined]{algorithm2e}
\usepackage{graphicx}
\usepackage{booktabs}
\usepackage{array}
\usepackage{enumitem}
\usepackage{amsthm}
\usepackage{algpseudocode}
\usepackage[utf8]{inputenc}
\usepackage{eqparbox}
\usepackage[nopar]{lipsum}
\usepackage{multirow}
\usepackage{makecell}
\usepackage{xcolor}
\usepackage{hhline} 
\usepackage{microtype}
\usepackage{diagbox}
\usepackage{longtable}
\usepackage{adjustbox}
\usepackage{mdframed}
\usepackage{framed,color} 
\usepackage{balance}

\definecolor{shadecolor}{rgb}{.92, .92, .92}

\usepackage{listings}
\usepackage{xcolor}
\usepackage{relsize}

\usepackage{booktabs,multirow,array}

\usepackage[many]{tcolorbox}
\newtcolorbox{fancyquotes}{%
    enhanced jigsaw, 
    breakable,      
    frame hidden,   
    left=0.5cm,       
    right=0.1cm,      
    overlay={%
        \node [scale=8,
            text=black,
            inner sep=0pt,] at ([xshift=-1cm,yshift=-1cm]frame.north west){}; 
        \node [scale=8,
            text=black,
            inner sep=0pt,] at ([xshift=1cm]frame.south east){};  
            },
                parbox=false,
}

\makeatletter

\usepackage{pifont} 
\definecolor{darkgreen}{rgb}{0.0,0.5,0.0}
\newcommand{\cmark}{\textcolor{darkgreen}{\ding{51}}}
\newcommand{\xmark}{\textcolor{red}{\ding{55}}}

\definecolor{c4}{RGB}{255,225,187}
\definecolor{c2}{RGB}{209, 233, 184}
\definecolor{c3}{RGB}{137 170 123}
\definecolor{c1}{RGB}{249, 229, 229}
\definecolor{c5}{RGB}{255, 128, 128}
\definecolor{c6}{RGB}{251, 132, 002}

\newtheorem*{proof*}{Proof}

\AtBeginDocument{%
  }

\copyrightyear{2025}
\acmYear{2025}
\setcopyright{rightsretained}
\acmConference[SIGIR '25]{Proceedings of the 48th International ACM SIGIR Conference on Research and Development in Information Retrieval}{July 13--18, 2025}{Padua, Italy}
\acmBooktitle{Proceedings of the 48th International ACM SIGIR Conference on Research and Development in Information Retrieval (SIGIR '25), July 13--18, 2025, Padua, Italy}
\acmDOI{10.1145/3726302.3730288}
\acmISBN{979-8-4007-1592-1/2025/07}

\begin{document}
\begin{sloppypar}
\title{MRAMG-Bench: A Comprehensive Benchmark for Advancing Multimodal Retrieval-Augmented Multimodal Generation}


\author{Qinhan Yu}
\authornote{These authors contributed equally to this work.}
\authornote{Work done during an internship at Huawei.}
\email{yuqinhan@stu.pku.edu.cn}
\orcid{0009-0004-0445-0786}
\affiliation{
    \institution{Peking University}
    \city{Beijing}
    \country{China}
}

\author{Zhiyou Xiao}
\authornotemark[1]
\email{xiaozhiyou@stu.pku.edu.cn}
\orcid{0009-0002-8919-0113}
\affiliation{
    \institution{Peking University}
    \city{Beijing}
    \country{China}
}

\author{Binghui Li}
\authornotemark[1]
\email{libinghui@pku.edu.cn}
\orcid{0009-0004-3506-3770}
\affiliation{
    \institution{Peking University}
    \city{Beijing}
    \country{China}
}

\author{Zhengren Wang}
\email{wzr@stu.pku.edu.cn}
\orcid{0000-0003-3541-9322}
\affiliation{
    \institution{Peking University}
    \city{Beijing}
    \country{China}
}

\author{Chong Chen}
\authornote{Corresponding Authors.}
\email{chenchong55@huawei.com}
\orcid{0000-0003-1417-2295}
\affiliation{
    \institution{Huawei Cloud BU}
    \city{Beijing}
    \country{China}
}

\author{Wentao Zhang}
\authornotemark[3]
\email{wentao.zhang@pku.edu.cn}
\orcid{0000-0002-7532-5550}
\affiliation{
    \institution{Peking University}
    \city{Beijing}
    \country{China}
}

\renewcommand{\shortauthors}{Yu and Xiao and Li et al.}

\begin{abstract}
  Recent advances in Retrieval-Augmented Generation (RAG) have significantly improved response accuracy and relevance by incorporating external knowledge into Large Language Models (LLMs). However, existing RAG methods primarily focus on generating text-only answers, even in Multimodal Retrieval-Augmented Generation (MRAG) scenarios, where multimodal elements are retrieved to assist in generating text answers. To address this, we introduce the Multimodal Retrieval-Augmented Multimodal Generation (MRAMG) task, in which we aim to generate multimodal answers that combine both text and images, fully leveraging the multimodal data within a corpus. Despite growing attention to this challenging task, a notable lack of a comprehensive benchmark persists for effectively evaluating its performance. To bridge this gap, we provide MRAMG-Bench, a meticulously curated, human-annotated benchmark comprising 4,346 documents, 14,190 images, and 4,800 QA pairs, distributed across six distinct datasets and spanning three domains: Web, Academia, and Lifestyle. The datasets incorporate diverse difficulty levels and complex multi-image scenarios, providing a robust foundation for evaluating the MRAMG task. To facilitate rigorous evaluation, MRAMG-Bench incorporates a comprehensive suite of both statistical and LLM-based metrics, enabling a thorough analysis of the performance of generative models in the MRAMG task. Additionally, we propose an efficient and flexible multimodal answer generation framework that can leverage LLMs/MLLMs to generate multimodal responses. Our datasets and complete evaluation results for 11 popular generative models are available at \url{https://github.com/MRAMG-Bench/MRAMG}.
  
  
\end{abstract}



\maketitle

\section{Introduction}
\label{sec:intro}

Retrieval-Augmented Generation (RAG) addresses the challenges of outdated knowledge and hallucinatory outputs in Large Language Models (LLMs) by incorporating external knowledge into the reasoning process, enhancing response accuracy and relevance  \citep{hallucination, gupta2024rag}. Traditional RAG methods focus on retrieving textual knowledge, limiting their ability to leverage multimodal information such as images and tables.

With the evolution of generative models, the shift from traditional LLMs to Multimodal Large Language Models (MLLMs) has facilitated a more comprehensive 
understanding by integrating both textual and visual inputs \citep{wang2024comprehensive,wang2024qwen2}. 

\begin{figure*}[t]
    \centering    \includegraphics[width=0.9\linewidth]{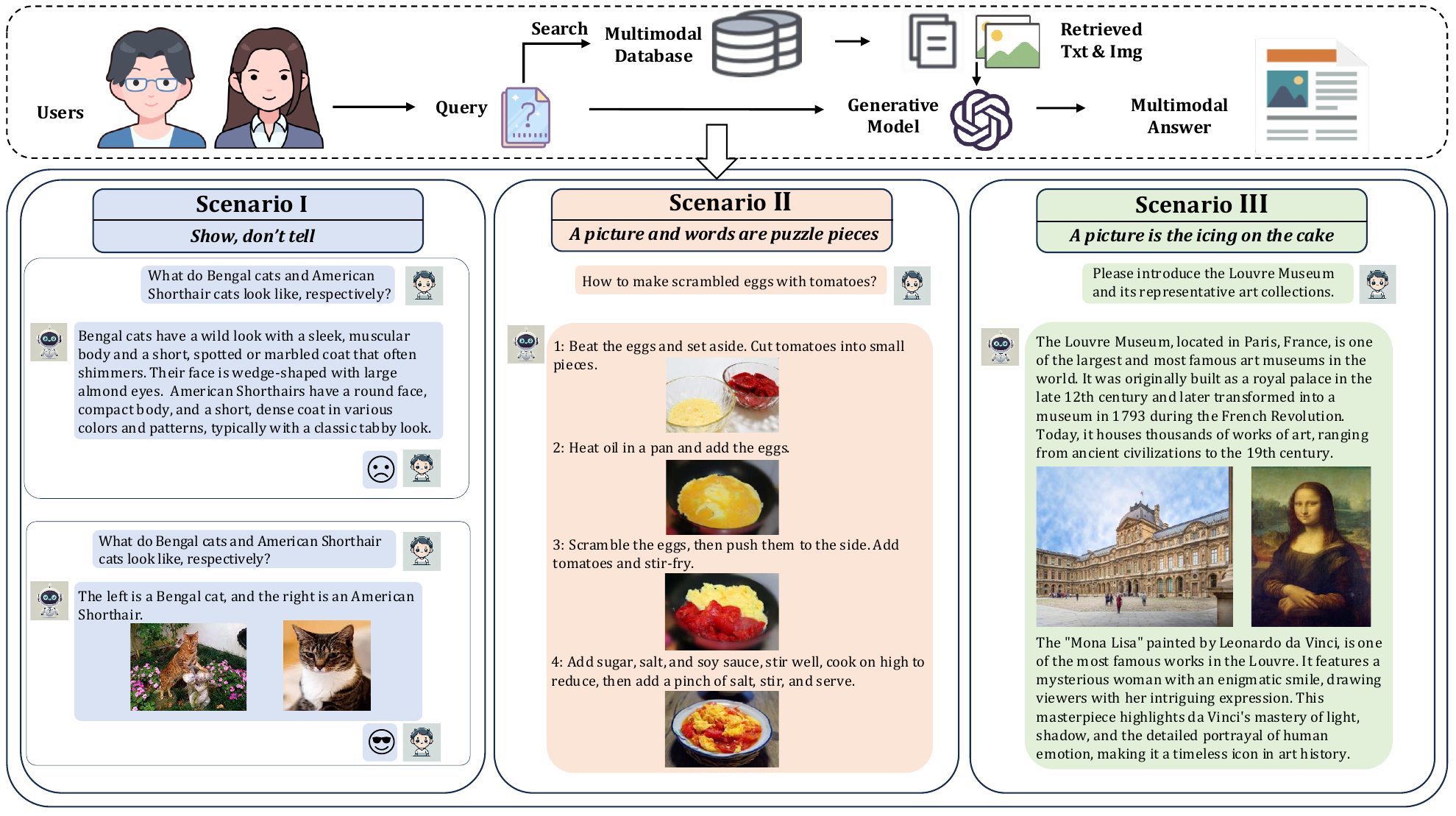}
   \vspace{-5mm}
   \caption{
    Illustration of the MRAMG task (above), with scenarios below showing how integrating text and images enhances clarity and understanding.
    }
    \label{fig:examples}
\end{figure*}
Correspondingly, Multimodal Retrieval-Augmented Generation (MRAG) \citep{chen2022murag} builds upon traditional RAG by incorporating both textual and visual information, thereby enhancing the quality of generated responses through the image understanding capabilities of MLLMs.
However, while the retrieval process incorporates multimodal information, current MRAG methods primarily focus on multimodal input and text-based output. 
Despite their advanced image comprehension capabilities, MLLMs frequently exhibit hallucinations when converting visual inputs into textual outputs \citep{bai2024hallucination}, which raises a natural question:
\begin{center}
    \small
    \textit{Why not present the images directly instead of describing their content?}
\end{center}

Indeed, there is an emerging trend in LLM deployment toward generating  multimodal outputs that seamlessly integrate both text and images.
Users increasingly favor visual content over purely textual information, preferring to ``see'' rather than merely be ``told''.
This growing need is particularly evident in critical scenarios, as shown in Figure \ref{fig:examples}:

\begin{itemize} 
    \item \textbf{Images as answers}: ``\textit{Show, don't tell}.'' In certain cases, an image provides the most effective response. For instance, when asked about the appearance of a cat, a photograph is the most direct and convincing answer—seeing truly is believing. 
    \item \textbf{Text-Image integration for 
    answers}: 
    ``\textit{A picture and words are two puzzle pieces—only together do they reveal the full story.}''
    Integrating pictures and words,  such as in step-by-step recipe instructions, significantly enhances comprehension which offers clearer insights and fosters a more intuitive understanding of the content.
    \item 
    \textbf{Images as a catalyst for richer answer}: 
    ``\textit{A picture is the icing on the cake.}''
    In applications such as tourist spot descriptions, integrating textual and visual content can significantly enhance the quality of the generated response.
    \end{itemize}
\vspace{-3pt}

To address this critical demand, we introduce the \textbf{Multimodal Retrieval-Augmented Multimodal Generation (MRAMG)} task, which aims to fully leverage the rich multimodal information within a corpus by generating integrated text-and-image answers (see Section \ref{sec:task_definition} for the formal definition).
However, current benchmarks for evaluating this critical task suffer from a notable lack of suitable datasets and scientifically rigorous evaluation metrics. 
This significant 
limitation hinders progress in this emerging field.

To bridge this gap, we introduce 
\textbf{MRAMG-Bench}, a novel benchmark designed to evaluate the MRAMG task comprehensively.
MRAMG-Bench consists of six 
carefully curated English datasets, comprising 4,346 documents, 14,190 images, and 4,800 QA pairs, 
sourced from three domains—Web, Academia, and Lifestyle—across seven distinct data sources.
Furthermore, MRAMG-Bench introduces unique challenges, such as hierarchical difficulty levels and order-based reasoning, which rigorously test the reasoning capabilities of both LLMs and MLLMs.  
To the best of our knowledge, MRAMG-Bench is the first benchmark that challenges models to autonomously determine the number, selection, and ordering of images in their responses, effectively simulating the complex scenarios encountered in real-world user interactions.

This benchmark is systematically constructed through a multi-stage workflow with both automatic construction of GPT-4o and meticulous formulation of human annotators, 
as illustrated in Figure \ref{fig:data_construction}.

Traditional evaluations of multimodal answers often rely on subjective metrics, such as image-text consistency, which are assessed through LLM-based judgments or human evaluations \citep{zhang2023internlm,zhu2024murar}.
However, these methods suffer from high subjectivity or evaluator inconsistency.
To mitigate these issues, we introduce a statistically grounded evaluation framework that systematically assesses both retrieval and generation performance. By integrating objective statistical metrics with LLM-based assessments, our framework delivers a comprehensive, multi-dimensional evaluation of multimodal answers, ensuring a fair and rigorous assessment process. Additionally, we also propose a general multimodal answer generation framework that integrates rule-based and model-based approaches. This flexible and scalable solution enables both LLMs and MLLMs to generate interleaved text-image responses.

Our main contributions can be summarized as follows:
\vspace{-1.5pt}
\begin{itemize}
    \item 
    We formulate the MRAMG task 
    and introduce MRAMG-Bench, a benchmark consisting of 4,800 human-annotated instances across multiple domains, featuring hierarchical difficulty levels and order-based reasoning challenges. 
    \item
    To ensure rigorous evaluation, we propose a robust, statistically grounded set of metrics that assess multimodal responses across various performance dimensions. 
    \item 
    We design an efficient and flexible generation framework that leverages LLMs/MLLMs to generate multimodal answers, combining rule-based and model-based approaches. 
    
    \item 
    We conduct a comprehensive evaluation of 11 advanced generative models on MRAMG-Bench, providing valuable insights into the capabilities and limitations of existing multimodal generation approaches. 
\end{itemize}
\vspace{-5pt}

\vspace{-3mm}
\section{Related Work}
\label{sec:related_work}

\subsection{Multimodal RAG}
Retrieval-Augmented Generation (RAG) \citep{2020RAG, zhao2024retrieval} enhances the capabilities of Large Language Models (LLMs) by integrating external knowledge, addressing limitations such as outdated training data and hallucinated outputs \citep{hallucination, gupta2024rag}. Recent advancements in Multimodal Large Language Models (MLLMs), which combine textual and visual inputs for generation tasks \citep{wang2024comprehensive}, have spurred the development of Multimodal Retrieval-Augmented Generation (MRAG) as an extension of traditional RAG. Notable approaches, including MuRAG \citep{chen2022murag} and REACT \citep{liu2023learning}, retrieve image-text pairs from external memory to support multimodal generation. In the context of Visual Question Answering (VQA), KAT \citep{gui2021kat} employs the CLIP \citep{radford2021learning} image encoder to associate specific image regions with external knowledge bases. Unlike existing MRAG methods that primarily generate textual outputs, our work focuses on a novel MRAMG task, where the output seamlessly integrates both textual and visual information. A closely related work, MuRAR \citep{zhu2024murar}, addresses source attribution by retrieving multimodal elements from attributed documents. M2RAG \citep{ma2024multi} further extends this by introducing a multi-stage image insertion framework, which involves multiple invocations of the generative model during a single answer generation process. However, these methods are often hindered by high computational costs due to repeated model calls. In contrast, we propose a more efficient framework for multimodal answer generation, leveraging a single invocation of the generative model.
\vspace{-3mm}
\subsection{RAG Benchmarks}
The effective evaluation of Retrieval-Augmented Generation (RAG) models is crucial for advancing their development and optimization. 
Established benchmarks \citep{MSMARCO,joshi2017triviaqa,yang2018hotpotqa,Naturalquestions,rajpurkar2016squad} are commonly used to evaluate RAG models \citep{petroni2020kilt, QAE}. MS-MARCO \citep{MSMARCO} is a large-scale question answering (QA) dataset based on real user queries. TriviaQA \citep{joshi2017triviaqa} consists of trivia questions that require evidence-based answers. HotpotQA \citep{yang2018hotpotqa} focuses on multi-hop reasoning. Natural Questions (NQ) \citep{Naturalquestions} is derived from real Google search queries. SQuAD \citep{rajpurkar2016squad} is a reading comprehension dataset with span-based answers.
While these text-based benchmarks have proven effective in assessing RAG performance, they fall short in evaluating multimodal tasks, which require the integration of both textual and visual information. 
To bridge this gap, we introduce a novel benchmark for evaluating the MRAMG task.
\vspace{-3mm}
\subsection{Multimodal RAG Benchmarks}
Similar to traditional RAG benchmarks, various multimodal RAG benchmarks \citep{marino2019ok,schwenk2022okvqa,jiang2024mmsearch,talmor2021multimodalqa,chang2022webqa,ma2024multi} have been developed to address tasks requiring multimodal knowledge. OK-VQA \citep{marino2019ok} and A-OKVQA \citep{schwenk2022okvqa} evaluate the multimodal reasoning ability of models using external knowledge beyond image content. MMSearch \citep{jiang2024mmsearch} evaluates MLLMs as multimodal search engines, focusing on image-to-image retrieval. MultiModalQA \citep{talmor2021multimodalqa} presents a more challenging scenario, where questions do not include images but require joint reasoning across text and tables to answer complex questions. While MultiModalQA relies on template-based questions, WebQA \citep{chang2022webqa} is a multi-hop, manually crafted multimodal QA dataset that involves retrieving relevant visual content for questions.
However, WebQA provides only textual answers, relies entirely on MLLMs for reasoning over retrieved images, and lacks textual support, making it unsuitable for language models that depend on linguistic context for coherent response generation. Notably, although M2RAG \citep{ma2024multi} constructs a multimodal corpus, it is limited to just 200 questions and a corpus confined to web pages. Additionally, the dataset relies on automated generation without manual verification and lacks ground truth for each query, further complicating accurate evaluation.
To address these challenges, our MRAMG-Bench introduces a comprehensive multimodal benchmark specifically designed for MRAMG task. Each question is meticulously paired with a precise, integrated text-image answer, enabling comprehensive and statistically rigorous evaluation.

\begin{table*}[tb!]
  \centering
   \caption{
Comparison of MRAMG-Bench with existing RAG benchmarks, where \includegraphics[height=1em]{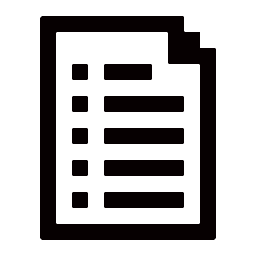} represents text and \includegraphics[height=1em]{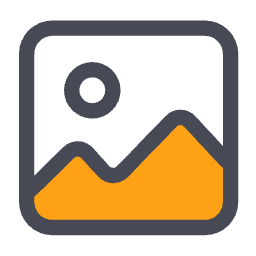} represents images.
   }
   \vspace{-3mm}
\resizebox{0.9\textwidth}{!}{
  \begin{tabular}{l|cc|cccc|ccc}
    \toprule
    \multirow{2}[4]{*}{Datasets} & \multicolumn{2}{c|}{Documents} & \multicolumn{4}{c|}{Questions}        & \multicolumn{3}{c}{Answers} \\
\cmidrule{2-10}          & Type  & Domain & Retrieval Modality & Difficulty Levels? & Type  & Num   & Exist? & Type  & Human-Annotated? \\
    \midrule
    HotpotQA \citep{yang2018hotpotqa} & \includegraphics[height=1em]{Fig/text.png}  & Web     & \includegraphics[height=1em]{Fig/text.png}  & \xmark    & \includegraphics[height=1em]{Fig/text.png}  & 113k   & \cmark    & \includegraphics[height=1em]{Fig/text.png}  & \cmark\\
    OK-VQA \citep{marino2019ok} & \includegraphics[height=1em]{Fig/text.png}  & Open Domain   & \includegraphics[height=1em]{Fig/text.png}  & \xmark   & \includegraphics[height=1em]{Fig/text.png} \includegraphics[height=1em]{Fig/pic.png} & 14k   & \cmark  & \includegraphics[height=1em]{Fig/text.png}  & \cmark\\
    WebQA \citep{chang2022webqa} & \includegraphics[height=1em]{Fig/text.png}  \includegraphics[height=1em]{Fig/pic.png} & Web      & \includegraphics[height=1em]{Fig/text.png} \includegraphics[height=1em]{Fig/pic.png} & \xmark    & \includegraphics[height=1em]{Fig/text.png}  & 56.6k & \cmark  & \includegraphics[height=1em]{Fig/text.png}  & \cmark\\
     MMSearch \citep{jiang2024mmsearch} &   \includegraphics[height=1em]{Fig/pic.png} & Multi-domain     &  \includegraphics[height=1em]{Fig/pic.png} & \xmark    & \includegraphics[height=1em]{Fig/text.png} \includegraphics[height=1em]{Fig/pic.png} & 300 & \cmark  & \includegraphics[height=1em]{Fig/text.png}  & \cmark\\
    M2RAG \citep{ma2024multi} & \includegraphics[height=1em]{Fig/text.png} \includegraphics[height=1em]{Fig/pic.png} & Web     & \includegraphics[height=1em]{Fig/text.png} \includegraphics[height=1em]{Fig/pic.png} & \xmark    & \includegraphics[height=1em]{Fig/text.png}  & 200   & \xmark    & \xmark    & \xmark \\
    \midrule
    MRAMG-Bench & \includegraphics[height=1em]{Fig/text.png} \includegraphics[height=1em]{Fig/pic.png} & Multi-domain    & \includegraphics[height=1em]{Fig/text.png} \includegraphics[height=1em]{Fig/pic.png} & \cmark  & \includegraphics[height=1em]{Fig/text.png}  & 4.8k   & \cmark  & \includegraphics[height=1em]{Fig/text.png} \includegraphics[height=1em]{Fig/pic.png} & \cmark\\
    \bottomrule
    \end{tabular}%
}
  \label{tab:compare}%
\end{table*}%

\vspace{-3mm}
\section{Task Formulation}
\label{sec:task_definition}

In this section, 
We formally define the task of
\textbf{Multimodal Retrieval-Augmented Multimodal Generation (MRAMG)}, focusing on constructing multimodal answers $\mathcal{A}$ based on a given text-form user query $q$ and its associated multimodal information $\mathcal{D}$. See Figure \ref{fig:examples} for an illustration.

\textbf{Multi-Document Multimodal Information.} We consider a multi-document multimodal knowledge base as the form $\mathcal{D} = \{d_1, d_2, \ldots, d_n\}$.
For each multimodal document $d\in \mathcal{D}$, it has the interleaved form $d = \{T_1, I_1, T_2, I_2, \dots, T_l, I_l, \dots \}$, where $T_i$ denotes the $i$-th text block (i.e., a semantic text paragraph) and $I_i$ denotes the $i$-th image within the document.

\textbf{Multimodal-Retrieve and Multimodal-Generation.} Specifically, the objective is to generate a multimodal answer $\mathcal{A} = \mathcal{F}(q, \mathcal{D}^*_q, \mathcal{M})$, where $\mathcal{F}$ represents the multimodal-answer generation method (i.e, generation framework), $\mathcal{D}^*_q = \{d_{j_1}, d_{j_2}, \dots, d_{j_k}\}$ 
represents the top-$k$ documents from $\mathcal{D}$ with the highest relevance to the query $q$, and $\mathcal{M}$ denotes a certain foundation generative model, including LLMs and MLLMs.
Indeed, given the user query $q$ and the retrieved documents $\mathcal{D}_q^*$ and the generative model $\mathcal{M}$, the generation framework $\mathcal{F}$ is required to return a multimodal answer $\mathcal{A}$ by selecting appropriate images from $\mathcal{D}_q^*$ and integrating them with the generated text.

\vspace{-3mm}
\section{Dataset Construction}
\label{sec:dataset}
\begin{figure*}[t]
    \centering
    \includegraphics[width=1.0\linewidth]{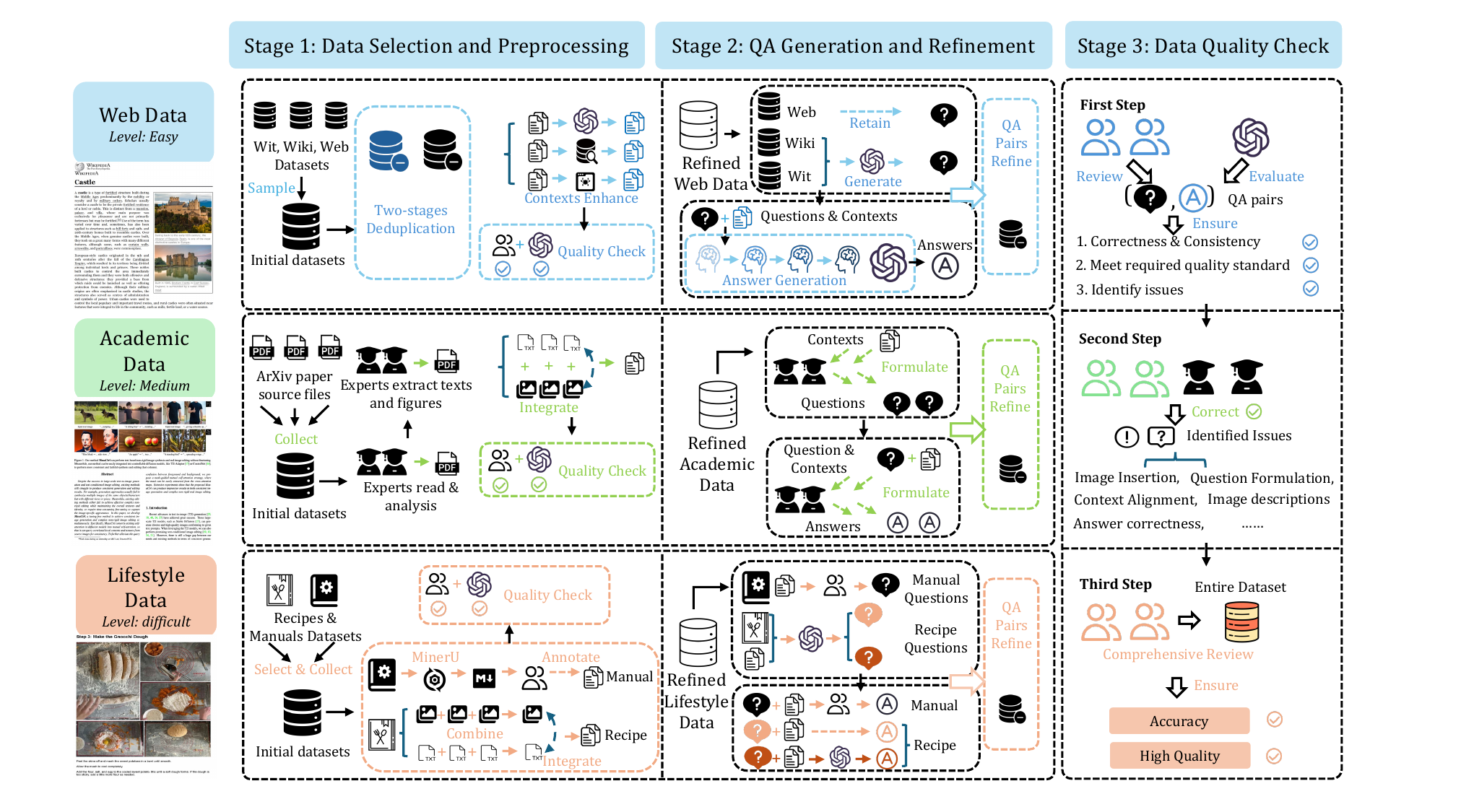}
    \vspace{-6mm}
    \caption{
    The MRAMG-Bench construction pipeline consists of three stages: (1) Data Selection and Preprocessing, where data is collected, cleaned, and preprocessed; (2) QA Generation and Refinement, involving the formulation and refinement of QA pairs; (3) Data Quality Check, where annotators and experts conduct a three-stage review to ensure high-quality benchmarks.
    }
    \label{fig:data_construction}
\end{figure*}

In this section, we present the construction details of MRAMG-Bench. It consists of three domains: Web, Academia and Lifestyle, which respectively contains web-based datasets—MRAMG-Wit, MRAMG-Wiki, MRAMG-Web, and 
an academic dataset-MRAMG-Arxiv, and two lifestyle datasets-MRAMG-Recipe, MRAMG-Manual. The process of creating datasets in different domains can be divided into three stages: (1) Data Selection and Preprocessing, (2) QA Generation and Refinement, (3) Data Quality Check. The overview of this process is shown in Figure \ref{fig:data_construction}.
\vspace{-3mm}
\subsection{Data Selection and Preprocessing}
We undertake three steps during this stage: (1) Data Collection, (2) Data Filtering and Context Enhancement, and (3) Data Refinement.

\subsubsection{Data Collection} The goal of this step is to collect MRAMG task-oriented multimodal document contextual data.
\vspace{-2mm}
\paragraph{Web Data.} This category includes Wikipedia pages and articles, derived from Wit dataset \citep{srinivasan2021wit}, WikiWeb2M dataset \citep{burns2023wikiweb2m}, and WebQA dataset \citep{chang2022webqa}. These datasets typically feature a rich integration of text and images. We select val sets and test sets from Wit dataset \citep{srinivasan2021wit}, and sample 82k entries from the 100k entries in the test set of WikiWeb2M dataset \citep{burns2023wikiweb2m} , and choose img-posFacts set in WebQA dataset \citep{chang2022webqa}, as the initial collection of MRAMG-Wit dataset, MRAMG-Wiki dataset and MRAMG-Web dataset.
These datasets primarily consist of single-image documents with relatively short length. Consequently, we categorize the three Web Data datasets as \textbf{level 1: easy} (difficulty level), where models can easily handle our task in this domain due to simple scenarios.
\vspace{-2mm}
\paragraph{Academic Data.} We collect 150 LaTeX source files and their corresponding PDF documents from the arXiv repository, with publication years spanning 2023 to 2024. These papers exhibit a seamless integration of text and images, providing an optimal foundation for tasks that demand precise image-text alignment. We manually select 110 high-quality papers with rich image-text interleaved content as the initial collection of MRAMG-Arxiv. It is defined as the \textbf{level 2: medium} dataset in our benchmark, where QA instances generally contain 1 to 4 images. Targeting at our task, it presents more challenges for models to retrieve and comprehend information from scientific document.
\vspace{-1mm}
\paragraph{Lifestyle Data.} Datasets in this domain feature in fine-integration of sequences of operation images and instruction texts. These datasets are thoroughly applicable to practical scenarios encountered in users' daily lives, which are collected from RecipeQA \citep{yagcioglu2018recipeqa}, ManualsLib\footnote{\url{https://www.manualslib.com}}
and the Technical Dataset\footnote{\url{https://www.kaggle.com/datasets/ahmedhgabr/technical-illustration/data}}.
We select the evaluation set from RecipeQA \citep{yagcioglu2018recipeqa}, and select 40 categories of instruction manuals with rich text-image content for subsequent construction of MRAMG-Recipe dataset and MRAMG-Manual dataset, respectively. We define Lifestyle Data as \textbf{level 3: difficult}, as they exhibit a significantly higher density of images and encompass diverse problem types, including single-image, no-image, and multi-image answers. 
Moreover, these datasets pose a significant challenge for models in determining the order of image insertion.
\subsubsection{Data Filtering and Context Enhancement}
The goal of this step is to clean data and acquire high-quality contexts, with images correctly inserted in the appropriate context positions.

For Web Data, we first filter out data entries with excessively long or short context, and data entries lacking images information or with overly large-sized images. We then implement two stages of deduplication: (1) MinHash-based deduplication \citep{shrivastava2014defense} method to remove data with similar images, and (2) string and semantic similarity measures \citep{TF-IDF,BGE} to further deduplicate data. Subsequently, we enhance the original contexts through crawling entire texts on original Wikipedia pages, and extending contexts leveraging GPT-4o based on the given information provided in the original data. For MRAMG-Web dataset, original collection contains questions where each question generally involves two entities and requires two distinct images for answering. We then extract entities included in the questions and retrieve entity-relevant texts in dataset as context supplements.

For Academic Data, we engage ten graduate
students specializing in computer science, with expertise in fields such as Natural Language Processing (NLP) and Computer Vision (CV), to serve as expert annotators. 
Each annotator is assigned papers within their specific domain of expertise, ensuring precise and contextually relevant annotations. 
The annotators meticulously read the selected papers, identify and extract textual content and corresponding figures that are most suitable for creating interleaved text-image contexts. This ensures accurate alignment of text and images in contexts.

For Lifestyle Data, in MRAMG-Recipe dataset, each recipe contains multiple step texts and each step text contains multiple images. We first combine multiple images into one image for each recipe step text and insert the combined image at the end of each step text. We then aggregate all the recipe step texts as an entire image-text interleaved recipe context. In MRAMG-Manual dataset, we employ MinerU \citep{wang2024mineru} to parse the PDF manuals into markdown format, preserving the original contextual order of images and texts within each manual.
A team of annotators is tasked with removing sensitive or irrelevant information, adjusting image placements within the content, and correcting any errors in the text.

\subsubsection{Data Refinement}
The final step in this stage is to refine image-text interleaved contexts along with the questions or answers existing in the original data. We leverage both GPT-4o and human verification for quality check. This step aims to achieve two key objectives: correct and consistent image-text context content and appropriate image positions in contexts.

\subsection{QA Generation and Refinement}
This stage can be further divided into three steps: (1) Question Generation, (2) Answer Generation, (3) QA Pair Refinement.

\subsubsection{Question Generation}
For MRAMG-Web dataset, we retain the questions contained in original dataset. As for MRAMG-Wiki, MRAMG-Wit and MRAMG-Recipe datasets, we leverage GPT-4o to generate high-quality, context-specific questions tailored to the provided text and images. Moreover, the questions in MRAMG-Manual are generated manually by human annotators. As for MRAMG-Arxiv dataset, we engage expert annotators to meticulously formulate questions. Overall, we adhere to the following two criteria: (1) The questions are constructed based on the corresponding contexts. (2) 
The questions should be natural and practical, with the potential to be effectively answered through the integration of text and images.

\subsubsection{Answer Generation} 
For Web Data, given the generated or original questions $Q$ and corresponding context $C$, we generate image-text interleaved answers $A$ using GPT-4o following a CoT reasoning strategy \citep{COT}: (1) \textbf{Question Validity Assessment}, we first ensure that valid questions $ValQ$ can be directly answered by contexts $C$ and remove invalid questions $InvalQ$. (2) \textbf{Evidence Extraction}, we then extract evidence $E$ from contexts $C$ to substantiate the answer $A$ for later answer construction. (3) \textbf{Answer Construction}, we construct a highly reliable, accurate and coherent answer based on valid questions $ValQ$ and evidence $E$. (4) \textbf{Image Integration} we finally enrich the constructed answers $A$ with a series of images integrated in the optimal positions. 
For Academic Data, expert annotators review existing questions, create answers incorporating images or tables, and integrate visual elements to ensure clear, multimodal comprehension of the content. For Lifestyle Data, in MRAMG-Recipe dataset, there are two types of questions: questions related to the entire recipe (recipe-specific) and questions only related to a specific step (step-specific). We combine all the step texts as the answers of recipe-specific questions and use answers generated by GPT-4o as the answers of step-specific questions. We manually insert images at the end of answer texts. In MRAMG-Manual dataset, answers are meticulously formulated by annotators, to ensure accurate selection and order of images in each manual.

\subsubsection{QA Pair Refinement}
We further refine the QA pairs formulated in the previous steps following the optimization approach outlined by \citet{zhu2024rageval}.
We leverage GPT-4o to extract supporting evidence from contexts and verify its alignment with keywords in the answer. We resolve the contradictions in the answer by cross-referencing the original answer and the extracted evidence, and discard irrelevant information to ensure the answer is contextually grounded.
\vspace{-3mm}
\subsection{Data Quality Check}

To ensure a high quality and reliability of MRAMG, we further check the consistency and correctness of QA pairs and their alignment with context-related images. We engage annotators in a structured data quality check process.
The process is carried out in three steps:
\begin{itemize}
    \item In the first step, a group of annotators review all QA pairs, assessing their correctness and consistency with both the context and image descriptions. They identify any problematic entries for further revision. Moreover, we also integrate GPT-4o for QA Evaluation, to further enhance the quality assurance process.
    \item In the second step, a separate group of annotators correct any identified issues, including those related to image insertions, question formulations, context alignment, image descriptions, and answer correctness. Specifically, expert annotators are further tasked with a meticulous review and correcting over the MRAMG-Arxiv dataset.
    \item In the third step, a team of 
    reviewers conduct a comprehensive recheck to ensure the overall accuracy of the dataset. This step validates that all corrections have been properly implemented and that the datasets meet the required quality standards.
\end{itemize}

\vspace{-3mm}
\subsection{Data Statistics}
\label{sec:statistics}
\begin{table*}[htbp]
  \centering
  \caption{The statistics of MRAMG-Bench. Multimodal Element Density refers to the proportion of images to text within a document, indicating image density relative to document length.}
  \vspace{-3mm}
    \resizebox{0.9\textwidth}{!}{
    \begin{tabular}{lccccccc}
    \toprule
    \textbf{Statistic} & \textbf{MRAMG-Wit} & \textbf{MRAMG-Wiki} & \textbf{MRAMG-Web} & \textbf{MRAMG-Arxiv} & \textbf{MRAMG-Recipe} & \textbf{MRAMG-Manual} & \textbf{Overall} \\
    \midrule
    \textbf{Difficulty Level} & 1     & 1     & 1     & 2     & 3     & 3     & 1, 2, 3 \\
    \textbf{Queries} & 600   & 500   & 750   & 200   & 2360  & 390   & 4800 \\
    \textbf{Documents} & 639   & 538   & 1500  & 101   & 1528  & 40    & 4346 \\
    \textbf{Images} & 639   & 538   & 1500  & 337   & 8569  & 2607  & 14190 \\
    \textbf{Avg Images Per Doc} & 1.00  & 1.00  & 1.00  & 3.34  & 5.61  & 65.18 & 3.27 \\
    \textbf{Avg Length Per Doc} & 532.31 & 865.28 & 98.42 & 853.5 & 485.22 & 6365.4 & 468.37 \\
    \textbf{Multimodal Element Density} & 1.88$\times 10^{-3}$ & 1.16$\times 10^{-3}$ & 1.02$\times 10^{-2}$ & 3.91$\times 10^{-3}$ & 1.16$\times 10^{-2}$ & 1.02$\times 10^{-2}$ & 6.98$\times 10^{-3}$ \\
    \textbf{Text-Only QA} & 0     & 0     & 0     & 0     & 92    & 40    & 132 \\
    \textbf{Single-Image QA} & 600   & 500   & 0     & 184   & 1481  & 92    & 2858 \\
    \textbf{Two-Image QA} & 0     & 0     & 750   & 12    & 60    & 126   & 948 \\
    \textbf{Three+ Images QA} & 0     & 0     & 0     & 3     & 727   & 132   & 862 \\
    \bottomrule
    \end{tabular}%
    }
  \label{tab:statistics}%
\end{table*}%

The MRAMG-Bench, summarized in Table~\ref{tab:statistics}, comprises 4,800 QA pairs across three distinct domains: 
Web, Academia, and Lifestyle. 
It integrates six datasets—\textbf{MRAMG-Wit}, \textbf{MRAMG-Wiki}, \textbf{MRAMG-Web}, \textbf{MRAMG-Arxiv}, \textbf{MRAMG-Recipe}, and \textbf{MRAMG-Manual}—containing 4,346 documents and 14,190 images, with tasks categorized into three difficulty levels. 
Datasets like MRAMG-Recipe and MRAMG-Manual feature longer texts and higher image density, posing greater challenges compared to others.
The benchmark's query distribution captures varying levels of complexity:
Single-image QA primarily involves direct image-text associations.
Two-image QA requires comparative analysis or step-wise reasoning.
Three+ image QA focuses on complex multi-step processes.
Text-only QA emphasizes pure textual reasoning and serves to evaluate the model's ability to exclude irrelevant images when they are not required.

\section{Generation Framework for MRAMG Task}
\label{sec:framework}

In this section, we present our generation framework, which consists of two stages: (1) the retrieval of relevant documents, and (2) the generation of multimodal answers based on the retrieved information, utilizing a foundational generative model. In the second stage, we propose three distinct strategies for answer generation: (a) LLM-based method: directly generating multimodal answers using LLMs, (b) MLLM-based method: directly generating multimodal answers using MLLMs, and (c) rule-based method: first obtaining a pure text answer from generative large models and then applying our matching-based algorithm to insert appropriate images into the text answer, resulting in multimodal answers. All the prompts we used can be found at \url{https://github.com/MRAMG-Bench/MRAMG}.

\subsection{Retrieving Relevant Documents}
We employ the BGE-M3 embedding model \citep{chen2024bge} to select the top-$k$ most relevant documents by calculating the cosine similarity between the user query $q$ and each document $d_i$ in the embedding space.
Then, we sequentially concatenate the text blocks and image blocks from each selected document, resulting in the multimodal information to be used: the text information $\mathcal{T}_q^* = \{t_1, t_2, \dots, t_l\}$ and the image infomation $\mathcal{I}_q^* = \{i_1, i_2, \dots, i_n\}$.

\subsection{Generating Multimodal Answer}


\subsubsection{LLM-Based Method}

We first consider how to generate multimodal answers using LLMs. Since LLMs cannot directly process image formats such as images for understanding, we choose an alternative method. 
In fact, we notice that the captions and surrounding context of images often offer detailed descriptions and supplementary information, which enhance the understanding of the images in the context of the query. Thus, we leverage the contextual information of images within the text context, as well as the captions of the images themselves, as textual substitute for the images when inputting into the language model. Specifically, we insert image placeholders $p_1, p_2, \dots, p_n$ at the corresponding positions of the retrieved text information $\mathcal{T}_q^*$, resulting in a new context $\mathcal{C}_q^* = \{t_1, t_2, \dots, t_{l_1}, p_1, t_{l_1 + 1}, \dots, t_{l_j}, p_j, t_{l_j +1}, \dots\}$. We consider that the context $t_{l_j}$ and $t_{l_j + 1}$ around the placeholder $p_j$ for the $j$-th image $i_j$ provide a suitable textual description for that image. Then, for a given LLM $\mathcal{M}$, we generate the multimodal answer as $\mathcal{A} = \mathcal{M}(\mathcal{P}_{llm}, q, \mathcal{C}_q^{*})$, where $\mathcal{P}_{llm}$ is the prompt, and the images in the output are represented as placeholders.

\subsubsection{MLLM-Based Method}

For MLLMs, we can directly input the textual information $\mathcal{T}_q^{*}$ and the image information $\mathcal{I}_q^{*}$. However, current open-source or closed-source MLLMs typically impose limits on the number of input images. Therefore, we are unable to use all the images in $\mathcal{I}_q^{*}$ as input. To address this, we propose leveraging the CLIP-based multimodal embedding model \citep{koukounas2024jina} to compute the embedding similarity between each image and the query, and then selecting the top-N images based on the similarity for input, denoted as $\operatorname{TopN}(q, \mathcal{I}_q^*)$. Formally, for a given MLLM $\mathcal{M}$, we generate the multimodal answer as $\mathcal{A} = \mathcal{M}(\mathcal{P}_{mllm}, \mathcal{T}_q^{*}, \operatorname{TopN}(q, \mathcal{I}_q^*))$, where $\mathcal{P}_{mllm}$ is the prompt, and the images in the output are also represented as placeholders.

\subsubsection{Rule-Based Method}

Here, we consider an alternative strategy for generating multimodal answers. Specifically, we first leverage the generative model to generate a pure textual answer based on the provided context. Then, we employ a matching-based insertion algorithm to select appropriate images from the retrieved set and insert them at suitable positions within the answer, thereby forming the multimodal answer. It is carried out in three steps:

\textbf{Step 1: Generate Text Answer.} We first use large generative model $\mathcal{M}$ to generate a pure text answer $\mathcal{A}_{txt}=\mathcal{M}(\mathcal{P}_{txt},\mathcal{T}_q^*)$, where $\mathcal{P}_{txt}$ denotes the text-answer-generating prompt. Subsequently, we segment the pure textual answer into sentences, i.e., $\mathcal{A}_{txt} = \{s_1, s_2, \dots, s_m\}$. 

\textbf{Step 2: Construct Bipartite Graph.} We then consider constructing the following bipartite graph $\mathcal{B} = (\mathcal{S}, \mathcal{I}, \mathcal{E})$, where $\mathcal{S} = \{s_1, s_2, \dots, s_m\}$ represents the set of representative vertices corresponding to the sentences in all pure text answers, and $\mathcal{I} = \{i_1, i_2, \dots, i_n\}$ represents the set of representative vertices corresponding to retrieved images. Here, we construct the edge set $\mathcal{E}$ by calculating the string similarity $\alpha_{k,l}$ (using BLEU method \citep{papineni2002bleu}) and semantic similarity $\beta_{k,l}$ (using BGE-M3 model \citep{chen2024bge}) for each sentence-image pair $(s_k, i_l)$. Specifically, to filter out irrelevant images, we define the edge set as $\mathcal{E} = \{(k, l) \mid \alpha_{k,l} > \alpha^* \text{ or } \beta_{k,l} > \beta^*\}$, where $\alpha^*, \beta^*$ are two hyper-parameters as thresholds, and each egde $(k,l)$ has the weight $w_{k,l} = \lambda\alpha_{k,l} + (1-\lambda)\beta_{k,l}$, where $\lambda\in [0,1]$ is a hyper-parameter.

\textbf{Step 3: Insert Images by Matching Algorithms.} Indeed, we aim to solve the following optimization problem:

\begin{equation}
\label{problem:matching}
    \operatornamewithlimits{max}_{M \subset \mathcal{E}} \sum_{(k,l) \in M} w_{k,l} ,
\end{equation}
where $M$ is a subset of the edge set $\mathcal{E}$. Specifically, we consider two types of constraints for $M$:

\begin{itemize}
    \item Each image can be inserted at most once, but multiple images are allowed to be inserted after the same sentence.
    \item Each image can be inserted at most once, and each sentence can have at most one image inserted after it.
\end{itemize}

For the first constraint, we can solve for the optimal solution to problem \eqref{problem:matching} as (Max Similarity Algorithm): 
\begin{equation}
    M^* = \left\{(k^*, l) \mid k^* = \operatornamewithlimits{argmax}_{(k,l) \in \mathcal{E}} w_{k,l}, \forall l \in [n] \right\}.
\end{equation}

For the second constraint, $M$ forms a matching of the bipartite graph $\mathcal{B}$. We use the standard Edmonds' Blossom algorithm \citep{Edmonds_1965}
to solve the weighted maximum bipartite graph matching problem in this case (Graph Matching Algorithm). 

\begin{figure}
    \centering
\includegraphics[width=0.95\linewidth]{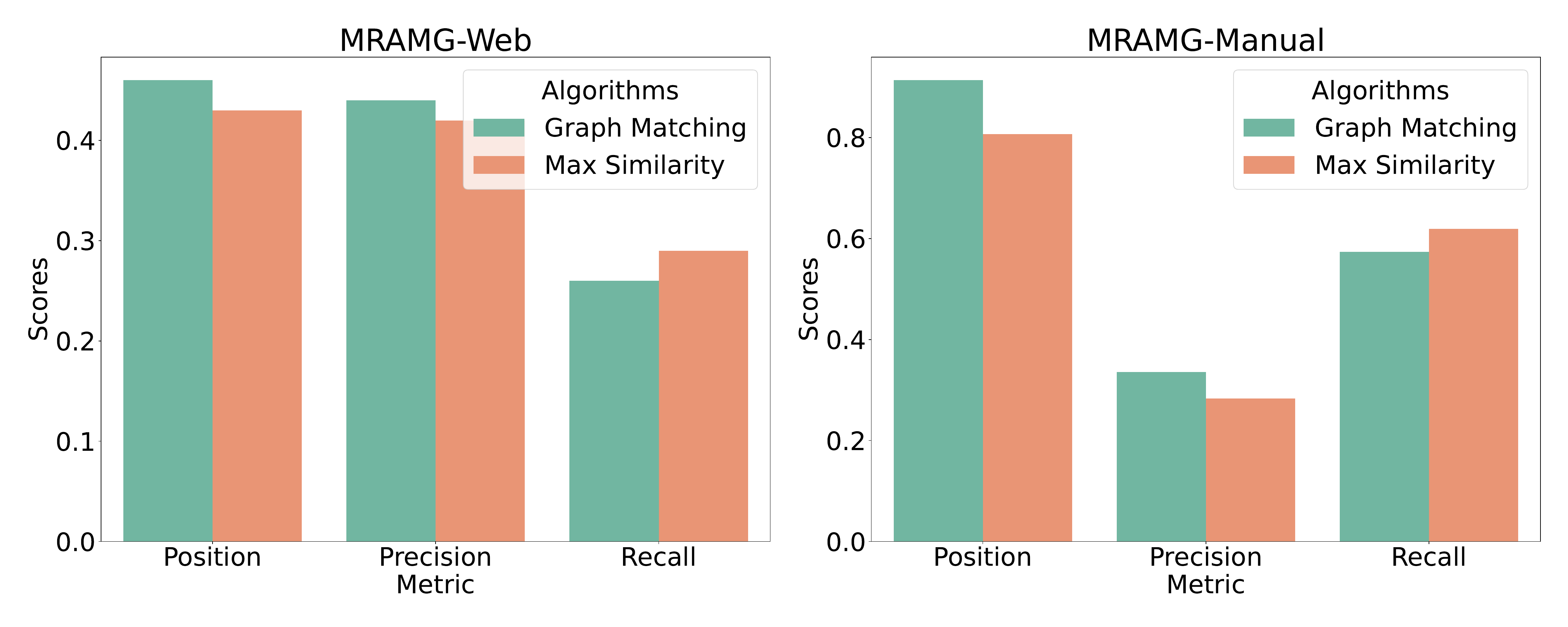}
\vspace{-3mm}
\caption{Comparison of the generative performance of two matching algorithms on different datasets.}
    \label{fig:enter-label}
\end{figure}

Finally, we insert the images into the output in the form of placeholders. We test the performances of the two matching-based insertion algorithms on the web and manual datasets, with results shown in Figure~\ref{fig:enter-label}.
Overall, the Graph Matching Algorithm, which restricts each sentence to have at most one image inserted after it, performs better. Therefore, we adopt this method for subsequent large-scale experimental evaluations.

\vspace{-5pt}
\section{Experiments}
\label{sec:experiments}

\subsection{Experimental Baselines and Settings}

\paragraph{Baselines} We evaluate 4 closed-source models and 7 open-source models. Specifically, we select 4 popular closed-source multimodal large models: GPT-4o \citep{gpt4o}, GPT-4o-mini \citep{gpt4o_mini}, Claude-3.5-Sonnet \citep{anthropic_claude_2024}, Gemini-1.5-Pro \citep{team2024gemini}. For the open-source models, we choose 4 multimodal large models: Qwen2-VL-7B-Instruct \citep{wang2024qwen2}, Qwen2-VL-72B-Instruct \citep{wang2024qwen2}, InternVL-2.5-8B \citep{chen2024expanding}, InternVL-2.5-78B \citep{chen2024expanding} and 3 text-based large models: 
DeepSeek-V3 \citep{liu2024deepseek}, Llama-3.1-8B-Instruct \citep{dubey2024llama} and Llama-3.3-70B-Instruct \citep{dubey2024llama}. Among these, for the four most widely used closed-source multimodal models, we test all the LLM/MLLM/rule-based methods in our framework. For the four other multimodal models, we test the MLLM/rule-based methods. For the remaining three text-based models, we test the LLM/rule-based methods.

\paragraph{Experiment Details} In the multimodal documents, images are replaced with <PIC> placeholders and each document is chunked using a size of 256 with the SentenceSplitter \citep{Liu_LlamaIndex_2022} to ensure efficient processing. During the retrieval phase, we use the BGE-M3 model \citep{chen2024bge} to retrieve the top-k $= 10$ relevant chunks, which contain the corresponding images. Concretely, images presented within the retrieved chunks are regarded as the corresponding retrieved images. In the evaluation stage, we utilize GPT-4o \citep{gpt4o} as the judging model to assess the performance of LLM-based metrics.

\begin{table}[tb!]
\vspace{-1mm}
\caption{Retrieval performance of the BGE-M3 model on 
MRAMG-Bench.}
  \centering
  \vspace{-3mm}
\resizebox{0.49\textwidth}{!}{
    \begin{tabular}{ccccccc}
    \toprule
    \textbf{Metrics} & \textbf{MRAMG-Wit} & \textbf{MRAMG-Wiki} & \textbf{MRAMG-Web} & \textbf{MRAMG-Arxiv} & \textbf{MRAMG-Recipe} & \textbf{MRAMG-Manual} \\
    \midrule
    \textbf{Visual Recall@10} & 0.99  & 0.99  & 0.99  & 0.94  & 0.87  & 0.72 \\
    \textbf{Context  Recall@10} & 0.9   & 0.89  & 0.95  & 0.83  & 0.91  & 0.87 \\
    \bottomrule
    \end{tabular}%

  \label{tab:retrieve}%
}
  \label{tab:latest1}%
\end{table}%

\begin{table*}[htbp]
  \centering
  \caption{Comprehensive performance results on MRAMG-Bench. Prec., Rec., B.S., Ord., Comp., Pos., and Avg. represent image precision, image recall, BERTScore,image ordering score, comprehensive score, image position score, and average score, respectively. The highest scores for each group are highlighted in bold.}
 \vspace{-3mm}
  \resizebox{\textwidth}{!}{
    \begin{tabular}{cl|cccccc|cccccc|ccccccc}
    \toprule
    \multirow{2}[4]{*}{\textbf{Framework}} & \multicolumn{1}{c|}{\multirow{2}[4]{*}{\textbf{Model}}} & \multicolumn{6}{c|}{\textbf{ Web Data}}       & \multicolumn{6}{c|}{\textbf{Academic Data}} & \multicolumn{7}{c}{\textbf{ Lifestyle Data}} \\
\cmidrule{3-21}          &       & \textbf{Prec.} & \textbf{Rec.} & \textbf{B.S.} & \textbf{Comp.} & \textbf{Pos.} & \textbf{Avg.} & \textbf{Prec.} & \textbf{Rec.} & \textbf{B.S.} & \textbf{Comp.} & \textbf{Pos.} & \textbf{Avg.} & \textbf{Prec.} & \textbf{Rec.} & \textbf{B.S.} & \textbf{Ord.} & \textbf{Comp.} & \textbf{Pos.} & \textbf{Avg.} \\
    \midrule
    \multicolumn{1}{c}{\multirow{11}[2]{*}{\textbf{Rule-Based}}} & \multicolumn{1}{p{4.125em}|}{GPT-4o } & 43.54 & 37.30 & 92.35 & 77.15 & 43.86 & 50.75 & 55.42 & 63.04 & 94.67 & 84.20 & 75.75 & 68.02 & 47.04 & 63.54 & 92.01 & 43.54 & 79.17 & \textbf{77.13} & \textbf{65.68} \\
          & GPT-4o-mini  & 38.20 & 33.08 & 91.10 & 76.59 & 38.57 & 46.87 & 51.71 & 59.29 & 94.36 & 85.20 & 73.75 & 66.07 & 49.14 & 61.52 & 91.13 & \textbf{43.87} & 79.32 & 75.89 & 64.92 \\
          & Claude-3.5-Sonnet  & 47.65 & 38.43 & 93.42 & \textbf{80.63} & 48.14 & 53.55 & 55.17 & 62.79 & 94.09 & 84.20 & 75.75 & 67.48 & \textbf{50.32} & 61.17 & 92.02 & 43.32 & \textbf{79.50} & 76.39 & 64.91 \\
          & Gemini-1.5-Pro  & 31.27 & 26.62 & 87.37 & 74.83 & 31.76 & 41.21 & 52.43 & 56.29 & 93.85 & 83.20 & 70.28 & 64.15 & 49.18 & 50.57 & 88.32 & 38.73 & 78.14 & 73.14 & 60.58 \\
          & DeepSeek-V3  & \textbf{55.49} & \textbf{46.62} & \textbf{94.12} & 80.18 & \textbf{56.16} & \textbf{59.04} & \textbf{56.12} & \textbf{67.29} & \textbf{94.90} & \textbf{84.30} & \textbf{78.46} & \textbf{70.05} & 27.07 & 57.56 & \textbf{92.42} & 24.07 & 74.19 & 65.19 & 57.22 \\
          & Qwen2-VL-7B-Instruct  & 37.98 & 34.81 & 91.26 & 70.99 & 38.28 & 46.32 & 49.17 & 52.17 & 92.09 & 78.90 & 67.08 & 60.88 & 43.79 & 60.93 & 91.49 & 39.52 & 77.70 & 77.33 & 63.87 \\
          & Qwen2-VL-72B-Instruct  & 34.81 & 31.49 & 88.70 & 71.89 & 35.14 & 43.75 & 45.42 & 48.71 & 92.39 & 79.60 & 65.42 & 59.50 & 31.82 & 49.30 & 89.88 & 25.51 & 72.92 & 70.94 & 56.45 \\
          & InternVL-2.5-8B  & 30.78 & 28.38 & 87.45 & 69.88 & 31.05 & 40.55 & 39.20 & 48.29 & 91.42 & 76.70 & 65.17 & 58.16 & 29.39 & 51.47 & 90.11 & 23.28 & 72.76 & 71.29 & 55.79 \\
          & InternVL-2.5-78B  & 38.79 & 34.49 & 90.97 & 74.82 & 39.06 & 47.04 & 52.21 & 62.00 & 94.51 & 85.40 & 75.38 & 67.57 & 22.61 & \textbf{67.71} & 92.19 & 19.41 & 74.95 & 56.55 & 56.26 \\
          & Llama-3.1-8B-Instruct  & 23.86 & 20.54 & 82.64 & 58.40 & 26.15 & 33.63 & 21.50 & 23.08 & 85.92 & 58.70 & 29.00 & 35.32 & 28.23 & 36.25 & 81.29 & 18.33 & 65.12 & 60.90 & 46.44 \\
          & Llama-3.3-70B-Instruct & 48.84 & 41.73 & 92.87 & 77.60 & 49.59 & 54.31 & 53.00 & 58.17 & 94.39 & 83.60 & 73.42 & 65.55 & 30.27 & 50.38 & 92.91 & 25.33 & 73.08 & 69.53 & 57.30 \\
    \midrule
    \multicolumn{1}{c}{\multirow{8}[2]{*}{\textbf{MLLM-Based}}} & GPT-4o  & 82.75 & 80.57 & \textbf{94.70} & 85.95 & 84.65 & 79.66 & \textbf{60.39} & 74.29 & \textbf{95.15} & 87.50 & \textbf{90.39} & \textbf{76.87} & \textbf{43.77} & 44.68 & \textbf{92.50} & \textbf{32.47} & 81.36 & \textbf{77.35} & 61.01 \\
          & GPT-4o-mini  & 69.98 & 86.08 & 94.59 & 80.58 & 72.93 & 76.29 & 36.17 & 74.79 & 95.08 & 83.20 & 74.66 & 68.62 & 29.34 & 47.71 & 91.71 & 21.95 & 77.19 & 69.95 & 56.32 \\
          & Claude-3.5-Sonnet  & \textbf{91.15} & \textbf{93.68} & 93.73 & \textbf{86.70} & \textbf{93.71} & \textbf{85.51} & 47.12 & \textbf{83.50} & 94.65 & \textbf{87.60} & 86.38 & 74.84 & 29.35 & 52.09 & 90.92 & 21.88 & 79.85 & 74.80 & 57.71 \\
          & Gemini-1.5-Pro  & 89.27 & 90.49 & 92.13 & 83.77 & 91.05 & 83.30 & 58.13 & 80.25 & 94.30 & 85.90 & 83.61 & 75.14 & 38.59 & \textbf{57.40} & 90.05 & 31.99 & \textbf{81.37} & 69.84 & \textbf{61.58} \\
          & Qwen2-VL-7B-Instruct  & 26.48 & 32.65 & 87.51 & 60.98 & 30.24 & 40.19 & 1.63  & 4.00  & 84.62 & 49.80 & 4.46  & 21.11 & 9.66  & 15.15 & 84.84 & 4.26  & 55.92 & 16.22 & 26.69 \\
          & Qwen2-VL-72B-Instruct  & 59.65 & 60.59 & 92.69 & 80.50 & 61.49 & 64.16 & 31.99 & 44.87 & 93.53 & 84.20 & 55.16 & 55.56 & 19.61 & 26.25 & 91.21 & 12.36 & 74.36 & 39.94 & 41.36 \\
          & InternVL-2.5-8B  & 52.71 & 65.86 & 91.89 & 72.78 & 61.17 & 63.30 & 12.22 & 27.87 & 83.72 & 58.10 & 28.49 & 35.34 & 22.19 & 37.94 & 89.41 & 14.54 & 73.99 & 60.11 & 48.22 \\
          & InternVL-2.5-78B & 75.50 & 76.27 & 93.98 & 84.11 & 78.68 & 74.96 & 36.62 & 55.00 & 94.47 & 84.80 & 64.11 & 61.01 & 21.43 & 29.09 & 91.11 & 13.53 & 75.43 & 51.99 & 45.23 \\
    \midrule
    \multicolumn{1}{c}{\multirow{7}[2]{*}{\textbf{LLM-Based}}} & GPT-4o  & 80.86 & 77.92 & \textbf{94.92} & 85.74 & 81.41 & 77.73 & \textbf{65.28} & 76.54 & \textbf{95.23} & 88.90 & 84.84 & 76.91 & 47.48 & 62.40 & 92.29 & 42.55 & 80.60 & 80.43 & 66.11 \\
          & GPT-4o-mini  & 69.58 & 90.95 & 94.35 & 81.29 & 70.96 & 76.91 & 37.69 & 83.33 & 95.01 & 84.50 & 69.07 & 69.91 & 44.36 & 38.27 & \textbf{92.69} & 31.16 & 83.24 & 54.96 & 53.33 \\
          & Claude-3.5-Sonnet  & 92.47 & 93.86 & 94.32 & \textbf{86.17} & 93.43 & 85.88 & 62.17 & \textbf{88.00} & 94.37 & \textbf{89.60} & \textbf{88.17} & \textbf{79.22} & 59.83 & 64.45 & 91.51 & 51.51 & \textbf{84.63} & \textbf{82.58} & 69.04 \\
          & Gemini-1.5-Pro  & \textbf{93.63} & 93.84 & 93.48 & 84.35 & \textbf{94.29} & 85.85 & 59.85 & 78.63 & 94.32 & 87.60 & 80.15 & 75.00 & \textbf{62.23} & \textbf{68.34} & 90.83 & \textbf{54.62} & 83.10 & 79.66 & \textbf{70.80} \\
          & DeepSeek-V3  & 93.08 & \textbf{95.51} & 94.68 & 85.64 & 93.98 & \textbf{86.45} & 46.57 & 81.13 & 94.70 & 87.50 & 70.01 & 72.53 & 45.71 & 67.56 & 91.78 & 40.35 & 82.64 & 77.03 & 66.04 \\
          & Llama-3.1-8B-Instruct  & 28.87 & 31.35 & 82.53 & 52.59 & 31.31 & 38.94 & 1.50  & 2.00  & 80.61 & 43.40 & 4.00  & 18.36 & 11.71 & 12.75 & 75.36 & 6.21  & 40.97 & 17.15 & 23.23 \\
          & Llama-3.3-70B-Instruct & 74.26 & 95.49 & 94.35 & 82.44 & 76.01 & 79.35 & 38.78 & 84.88 & 95.01 & 83.40 & 64.59 & 68.93 & 35.29 & 69.34 & 91.89 & 30.60 & 80.15 & 70.66 & 61.86 \\
    \bottomrule
    \end{tabular}%
    }
  \label{table:main_results}%
\end{table*}%
 \vspace{-1mm}

\subsection{Evaluation Metric}

\subsubsection{Retrieval Evaluation}

To evaluate the retrieval performance, 
We consider the following metrics:

\begin{itemize}
    \item 
    \textbf{Context Recall \citep{es2023ragas}} uses LLMs to evaluate whether the retrieved documents contain all the relevant  textual information required for answer generation.
    \item 
    \textbf{Visual Recall} measures the percentage of retrieved images relative to the total number of images in the ground truth.
\end{itemize}

\subsubsection{Generation Evaluation}

To evaluate performance of multimodal answers, we consider the following metrics\footnote{See detailed metrics at 
\url{https://github.com/MRAMG-Bench/MRAMG}
.}, which can be divided into two categories: statistical-based metrics (first six metrics) and LLM-based metrics (last four metrics). 

\begin{itemize}
    \item 
    \textbf{Image Precision} measures the percentage of correct images in the multimodal answer relative to the total number of inserted images, assessing whether irrelevant images were introduced. 

    \item 
    \textbf{Image Recall} measures the percentage of correct images in the multimodal answer relative to the total number of images in the ground truth, evaluating whether the answer effectively includes useful image information. 

    \item 
    \textbf{Image F1 Score} is the harmonic mean of Precision and Recall, providing an overall evaluation of the image quality in the multimodal answer. 
    \item 
    \textbf{Image Ordering Score} evaluates whether the order of images inserted into the multimodal answer matches the order of images in the ground truth \footnote{Here, we apply the ordering score only to Lifestyle Data 
    , as instances in these datasets typically contain a large number of images, and the order of the images is crucial.}. Specifically, we compute the weighted edit distance between the two image sequences. 
    \item 
    \textbf{Rouge-L \citep{lin-2004-rouge}} is a text generation evaluation metric based on the longest common subsequence, measuring the structural similarity between the answer and the ground truth.
    \item 
    \textbf{BERTScore \citep{zhang2019bertscore}} is a text generation evaluation metric based on BERT \citep{devlin2018bert}, used to assess the semantic similarity between the text in the answer and the ground truth.
    \item 
    \textbf{Image Relevance} measures the relevance of the inserted image to the query-answer pair, specifically assessing whether the content described by the image is meaningfully related to the content of the QA \citep{zhu2024murar,ma2024multi}. 
    \item 
    \textbf{Image Effectiveness} measures the effectiveness of the images inserted into the multimodal answer, assessing whether the images align with the QA content and contribute to the understanding of the answer \citep{zhu2024murar,ma2024multi}. 
    \item 
    \textbf{Image Position Score} is used to assess the appropriateness of the image placement in the multimodal answer. 
    \item 
    \textbf{Comprehensive Score} reflects overall quality of the multimodal answer, evaluating whether the answer appropriately addresses the query and maintains overall coherence. 
\end{itemize}

\subsection{Experiment Results}

\subsubsection{Retrieval Performance}

As shown in Table \ref{tab:retrieve}, we observe that both visual recall and context recall perform well for Web Data 
, with MRAMG-Web achieving particularly high scores of 0.99 and 0.95, respectively. 
In contrast, both metrics progressively decline for the three other datasets
, which can be attributed to their longer text passages and higher image counts. Nevertheless, overall retrieval effectiveness remains strong. 
\vspace{-8pt}
\subsubsection{Generation Performance}
\label{generation}

In this section, we evaluate generation performance, with results presented for three domains (Web/ Academia/ Lifestyle). Due to space limitations, 
complete results are available at 
\url{https://github.com/MRAMG-Bench/MRAMG}.

\textbf{Generation Performance Across Different Datasets.} From the results in Table \ref{table:main_results}, it is evident that the overall generation performance significantly decreases as the dataset complexity increases, which aligns with our expectations regarding the varying difficulty of different datasets. Notably, the performance of rule-based method on simpler datasets is basically suboptimal, which can be attributed to data characteristics, particularly related to the Avg Images Per Doc metric. According to Table \ref{tab:statistics}, this metric is 1 for all the 
web datasets 
, whereas in the other datasets, it is substantially higher. With fewer images per document, the rule-based method struggles to effectively associate images with sentences, as multiple sentences often correspond to a single image. This results in ambiguous correlations and complicated threshold selection, leading to a higher error rate. In contrast, as the number of images increases, the gap in correlation between rule-based method and other methods becomes more pronounced. Under these conditions, rule-based performs better due to improved alignment with sentence-image relationships.

\textbf{Generation Performance Across Different Models.} As shown in Table \ref{table:main_results}, advanced models such as Gemini, Claude, GPT-4o, and Deepseek-V3 consistently outperform smaller open-source models ($\sim7$B parameters) across all domains and methods. These smaller models exhibit subpar performance across different methods and dataset domains, even when utilizing rule-based generation techniques. In contrast, larger open-source models ($\sim70$B parameters) significantly reduce the performance gap with closed-source models, achieving results within approximately a tenth of the margin on simpler datasets, such as Web Data. For example, InternVL-2.5-78B and Llama-3.3-70B-Instruct attain Image Precision scores of 75.5 and 74.26, respectively, with average scores surpassing 70. 

On more challenging datasets, however, the performance gap becomes more pronounced, highlighting the limitations of open-source models in handling complex MRAMG tasks. Specifically, when employing MLLM-based method, Qwen2-VL-72B-Instruct achieves an Image Recall score of only 26.25 on Lifestyle Data, which is significantly lower than Gemini's 57.4. This result underscores the challenges faced by even 70B-scale models in accurately identifying images in complex scenarios.  
Nevertheless, smaller open-source models remain a cost-effective solution for simpler applications with limited computational resources. 

In particular, the order metric for the lifestyle domain demonstrates poor performance across all models and methods, with none achieving a passing score. Notably, GPT-4o performs best under both the rule-based and MLLM-based methods, scoring 43.54 and 32.47, respectively, while the Gemini model attains the highest score of 54.62 under the LLM-based method. However, even these scores fall short of the passing threshold, underscoring the ambiguity in LLM and MLLM models regarding the concept of image insertion order, which remains an unresolved challenge.

\textbf{Generation Performance Across Different Generation Methods.}  
In comparing different methods, an overall performance trend emerges: $\text{LLM-based} > \text{MLLM-based} > \text{Rule-based}$.  
For example, on Web Data, the Gemini model achieves average scores of 85.85, 83.3, and 41.21 for LLM-based, MLLM-based, and rule-based methods, respectively.

\textbf{LLM-Based Method:} By integrating contextual information surrounding images into the generation process, this method achieves natural and precise image insertion, underscoring the critical role of context in ensuring insertion accuracy.

\textbf{MLLM-Based Method:} While effective on simpler datasets such as Web Data, their performance deteriorates on more challenging datasets due to the increased difficulty of distinguishing visually similar images, revealing the current limitations of model capabilities. In this category, the Gemini model achieves the highest average scores on Web and Lifestyle Data, scoring 85.51 and 61.58, respectively, while Claude outperforms others on Academic Data with a score of 76.87.

\textbf{Rule-Based Method:} Although these methods exhibit significantly lower performance compared to model-based approaches on simpler datasets, the performance gap diminishes as dataset complexity increases. 
In the lifestyle domain,
rule-based methods even outperform certain model-based approaches. For instance, GPT-4o achieves an average score of 65.68 using rule-based method, surpassing its MLLM-based score of 61.01. Notably, the Deepseek-V3 model demonstrates outstanding performance on Web Data and Academic Data, achieving top scores across almost all metrics. This suggests that for simpler datasets, rule-based method aligns well with the Deepseek-V3 model, indicating the model's tendency to leverage such approaches effectively. Despite their limitations, rule-based methods offer several key advantages:  
\vspace{-2pt}
\begin{itemize}
    \item \textbf{Flexibility:} Not constrained by context window sizes or input image limits.
    \item \textbf{Cost Efficiency:} Reduces computational costs by up to one-third compared to LLM-based methods and by half compared to MLLM-based methods.
    \item \textbf{Stability:} Avoids instability issues such as erroneous placeholder generation.
\end{itemize}
\vspace{-2pt}
Overall, rule-based methods provide a viable and efficient alternative, particularly in resource-constrained or stability-critical scenarios. Furthermore, the performance of LLM-based methods is generally on par with or even superior to rule-based methods, showcasing the strong \textit{in-context reasoning} capabilities of large models.

\vspace{-7.5pt}
\section{Conclusion}
\label{sec:conclusion}
Driven by the growing demand for multimodal answers, MRAMG has emerged as a critical task that aligns with real-world user needs.
To address the lack of evaluation resources for this task, we present MRAMG-Bench, a curated benchmark containing 4,800 QA pairs across diverse domains and varying difficulty levels.
We further propose a comprehensive evaluation strategy that incorporates statistical and LLM-based metrics to rigorously assess both retrieval and generation performance.
Additionally, we introduce a general MRAMG framework to enable models to produce interleaved text-image responses. Our evaluation of 11 popular generative models highlights notable limitations in handling challenging datasets and selecting the correct image order, emphasizing the necessity for deeper exploration of the MRAMG task.

\vspace{-7.5pt}
\begin{acks}
    This work is supported by National Natural Science Foundation
of China (92470121, 62402016), National Key R\&D
Program of China (2024YFA1014003), and High-performance 
Computing Platform of Peking University. Binghui Li is supported by the Elite Ph.D. Program in Applied Mathematics for PhD Candidates in Peking University.
\end{acks}

\bibliographystyle{ACM-Reference-Format}
\balance
\bibliography{sample-base}


\begin{thebibliography}{51}


\ifx \showCODEN    \undefined \def \showCODEN     #1{\unskip}     \fi
\ifx \showISBNx    \undefined \def \showISBNx     #1{\unskip}     \fi
\ifx \showISBNxiii \undefined \def \showISBNxiii  #1{\unskip}     \fi
\ifx \showISSN     \undefined \def \showISSN      #1{\unskip}     \fi
\ifx \showLCCN     \undefined \def \showLCCN      #1{\unskip}     \fi
\ifx \shownote     \undefined \def \shownote      #1{#1}          \fi
\ifx \showarticletitle \undefined \def \showarticletitle #1{#1}   \fi
\ifx \showURL      \undefined \def \showURL       {\relax}        \fi
\providecommand\bibfield[2]{#2}
\providecommand\bibinfo[2]{#2}
\providecommand\natexlab[1]{#1}
\providecommand\showeprint[2][]{arXiv:#2}

\bibitem[Anthropic(2024)]%
        {anthropic_claude_2024}
\bibfield{author}{\bibinfo{person}{Anthropic}.} \bibinfo{year}{2024}\natexlab{}.
\newblock \bibinfo{title}{Claude 3.5 Sonnet}.
\newblock \bibinfo{howpublished}{\url{https://www.anthropic.com/news/claude-3-5-sonnet}}.
\newblock


\bibitem[Bai et~al\mbox{.}(2024)]%
        {bai2024hallucination}
\bibfield{author}{\bibinfo{person}{Zechen Bai}, \bibinfo{person}{Pichao Wang}, \bibinfo{person}{Tianjun Xiao}, \bibinfo{person}{Tong He}, \bibinfo{person}{Zongbo Han}, \bibinfo{person}{Zheng Zhang}, {and} \bibinfo{person}{Mike~Zheng Shou}.} \bibinfo{year}{2024}\natexlab{}.
\newblock \showarticletitle{Hallucination of multimodal large language models: A survey}.
\newblock \bibinfo{journal}{\emph{arXiv preprint arXiv:2404.18930}} (\bibinfo{year}{2024}).
\newblock


\bibitem[Burns et~al\mbox{.}(2023)]%
        {burns2023wikiweb2m}
\bibfield{author}{\bibinfo{person}{Andrea Burns}, \bibinfo{person}{Krishna Srinivasan}, \bibinfo{person}{Joshua Ainslie}, \bibinfo{person}{Geoff Brown}, \bibinfo{person}{Bryan~A Plummer}, \bibinfo{person}{Kate Saenko}, \bibinfo{person}{Jianmo Ni}, {and} \bibinfo{person}{Mandy Guo}.} \bibinfo{year}{2023}\natexlab{}.
\newblock \showarticletitle{Wikiweb2m: A page-level multimodal wikipedia dataset}.
\newblock \bibinfo{journal}{\emph{arXiv preprint arXiv:2305.05432}} (\bibinfo{year}{2023}).
\newblock


\bibitem[Chang and Bisk(2022)]%
        {chang2022webqa}
\bibfield{author}{\bibinfo{person}{Yingshan Chang} {and} \bibinfo{person}{Yonatan Bisk}.} \bibinfo{year}{2022}\natexlab{}.
\newblock \showarticletitle{WebQA: A Multimodal Multihop NeurIPS Challenge}. In \bibinfo{booktitle}{\emph{NeurIPS 2021 Competitions and Demonstrations Track}}. PMLR, \bibinfo{pages}{232--245}.
\newblock


\bibitem[Chen et~al\mbox{.}(2024b)]%
        {chen2024bge}
\bibfield{author}{\bibinfo{person}{Jianlv Chen}, \bibinfo{person}{Shitao Xiao}, \bibinfo{person}{Peitian Zhang}, \bibinfo{person}{Kun Luo}, \bibinfo{person}{Defu Lian}, {and} \bibinfo{person}{Zheng Liu}.} \bibinfo{year}{2024}\natexlab{b}.
\newblock \showarticletitle{Bge m3-embedding: Multi-lingual, multi-functionality, multi-granularity text embeddings through self-knowledge distillation}.
\newblock \bibinfo{journal}{\emph{arXiv preprint arXiv:2402.03216}} (\bibinfo{year}{2024}).
\newblock


\bibitem[Chen et~al\mbox{.}(2022)]%
        {chen2022murag}
\bibfield{author}{\bibinfo{person}{Wenhu Chen}, \bibinfo{person}{Hexiang Hu}, \bibinfo{person}{Xi Chen}, \bibinfo{person}{Pat Verga}, {and} \bibinfo{person}{William~W Cohen}.} \bibinfo{year}{2022}\natexlab{}.
\newblock \showarticletitle{Murag: Multimodal retrieval-augmented generator for open question answering over images and text}.
\newblock \bibinfo{journal}{\emph{arXiv preprint arXiv:2210.02928}} (\bibinfo{year}{2022}).
\newblock


\bibitem[Chen et~al\mbox{.}(2024a)]%
        {chen2024expanding}
\bibfield{author}{\bibinfo{person}{Zhe Chen}, \bibinfo{person}{Weiyun Wang}, \bibinfo{person}{Yue Cao}, \bibinfo{person}{Yangzhou Liu}, \bibinfo{person}{Zhangwei Gao}, \bibinfo{person}{Erfei Cui}, \bibinfo{person}{Jinguo Zhu}, \bibinfo{person}{Shenglong Ye}, \bibinfo{person}{Hao Tian}, \bibinfo{person}{Zhaoyang Liu}, {et~al\mbox{.}}} \bibinfo{year}{2024}\natexlab{a}.
\newblock \showarticletitle{Expanding performance boundaries of open-source multimodal models with model, data, and test-time scaling}.
\newblock \bibinfo{journal}{\emph{arXiv preprint arXiv:2412.05271}} (\bibinfo{year}{2024}).
\newblock


\bibitem[Devlin(2018)]%
        {devlin2018bert}
\bibfield{author}{\bibinfo{person}{Jacob Devlin}.} \bibinfo{year}{2018}\natexlab{}.
\newblock \showarticletitle{Bert: Pre-training of deep bidirectional transformers for language understanding}.
\newblock \bibinfo{journal}{\emph{arXiv preprint arXiv:1810.04805}} (\bibinfo{year}{2018}).
\newblock


\bibitem[Dubey et~al\mbox{.}(2024)]%
        {dubey2024llama}
\bibfield{author}{\bibinfo{person}{Abhimanyu Dubey}, \bibinfo{person}{Abhinav Jauhri}, \bibinfo{person}{Abhinav Pandey}, \bibinfo{person}{Abhishek Kadian}, \bibinfo{person}{Ahmad Al-Dahle}, \bibinfo{person}{Aiesha Letman}, \bibinfo{person}{Akhil Mathur}, \bibinfo{person}{Alan Schelten}, \bibinfo{person}{Amy Yang}, \bibinfo{person}{Angela Fan}, {et~al\mbox{.}}} \bibinfo{year}{2024}\natexlab{}.
\newblock \showarticletitle{The llama 3 herd of models}.
\newblock \bibinfo{journal}{\emph{arXiv preprint arXiv:2407.21783}} (\bibinfo{year}{2024}).
\newblock


\bibitem[Edmonds(1965)]%
        {Edmonds_1965}
\bibfield{author}{\bibinfo{person}{Jack Edmonds}.} \bibinfo{year}{1965}\natexlab{}.
\newblock \showarticletitle{Paths, Trees, and Flowers}.
\newblock \bibinfo{journal}{\emph{Canadian Journal of Mathematics}}  \bibinfo{volume}{17} (\bibinfo{year}{1965}), \bibinfo{pages}{449–467}.
\newblock
\href{https://doi.org/10.4153/CJM-1965-045-4}{doi:\nolinkurl{10.4153/CJM-1965-045-4}}


\bibitem[Es et~al\mbox{.}(2023)]%
        {es2023ragas}
\bibfield{author}{\bibinfo{person}{Shahul Es}, \bibinfo{person}{Jithin James}, \bibinfo{person}{Luis Espinosa-Anke}, {and} \bibinfo{person}{Steven Schockaert}.} \bibinfo{year}{2023}\natexlab{}.
\newblock \showarticletitle{Ragas: Automated evaluation of retrieval augmented generation}.
\newblock \bibinfo{journal}{\emph{arXiv preprint arXiv:2309.15217}} (\bibinfo{year}{2023}).
\newblock


\bibitem[Gui et~al\mbox{.}(2021)]%
        {gui2021kat}
\bibfield{author}{\bibinfo{person}{Liangke Gui}, \bibinfo{person}{Borui Wang}, \bibinfo{person}{Qiuyuan Huang}, \bibinfo{person}{Alex Hauptmann}, \bibinfo{person}{Yonatan Bisk}, {and} \bibinfo{person}{Jianfeng Gao}.} \bibinfo{year}{2021}\natexlab{}.
\newblock \showarticletitle{Kat: A knowledge augmented transformer for vision-and-language}.
\newblock \bibinfo{journal}{\emph{arXiv preprint arXiv:2112.08614}} (\bibinfo{year}{2021}).
\newblock


\bibitem[Gupta et~al\mbox{.}(2024)]%
        {gupta2024rag}
\bibfield{author}{\bibinfo{person}{Aman Gupta}, \bibinfo{person}{Anup Shirgaonkar}, \bibinfo{person}{Angels de~Luis Balaguer}, \bibinfo{person}{Bruno Silva}, \bibinfo{person}{Daniel Holstein}, \bibinfo{person}{Dawei Li}, \bibinfo{person}{Jennifer Marsman}, \bibinfo{person}{Leonardo~O Nunes}, \bibinfo{person}{Mahsa Rouzbahman}, \bibinfo{person}{Morris Sharp}, {et~al\mbox{.}}} \bibinfo{year}{2024}\natexlab{}.
\newblock \showarticletitle{RAG vs Fine-tuning: Pipelines, Tradeoffs, and a Case Study on Agriculture}.
\newblock \bibinfo{journal}{\emph{arXiv preprint arXiv:2401.08406}} (\bibinfo{year}{2024}).
\newblock


\bibitem[Huang et~al\mbox{.}(2023)]%
        {hallucination}
\bibfield{author}{\bibinfo{person}{Lei Huang}, \bibinfo{person}{Weijiang Yu}, \bibinfo{person}{Weitao Ma}, \bibinfo{person}{Weihong Zhong}, \bibinfo{person}{Zhangyin Feng}, \bibinfo{person}{Haotian Wang}, \bibinfo{person}{Qianglong Chen}, \bibinfo{person}{Weihua Peng}, \bibinfo{person}{Xiaocheng Feng}, \bibinfo{person}{Bing Qin}, {et~al\mbox{.}}} \bibinfo{year}{2023}\natexlab{}.
\newblock \showarticletitle{A survey on hallucination in large language models: Principles, taxonomy, challenges, and open questions}.
\newblock \bibinfo{journal}{\emph{arXiv preprint arXiv:2311.05232}} (\bibinfo{year}{2023}).
\newblock


\bibitem[Jiang et~al\mbox{.}(2024)]%
        {jiang2024mmsearch}
\bibfield{author}{\bibinfo{person}{Dongzhi Jiang}, \bibinfo{person}{Renrui Zhang}, \bibinfo{person}{Ziyu Guo}, \bibinfo{person}{Yanmin Wu}, \bibinfo{person}{Jiayi Lei}, \bibinfo{person}{Pengshuo Qiu}, \bibinfo{person}{Pan Lu}, \bibinfo{person}{Zehui Chen}, \bibinfo{person}{Guanglu Song}, \bibinfo{person}{Peng Gao}, {et~al\mbox{.}}} \bibinfo{year}{2024}\natexlab{}.
\newblock \showarticletitle{Mmsearch: Benchmarking the potential of large models as multi-modal search engines}.
\newblock \bibinfo{journal}{\emph{arXiv preprint arXiv:2409.12959}} (\bibinfo{year}{2024}).
\newblock


\bibitem[Jones(1973)]%
        {TF-IDF}
\bibfield{author}{\bibinfo{person}{Karen~Sparck Jones}.} \bibinfo{year}{1973}\natexlab{}.
\newblock \showarticletitle{Index term weighting}.
\newblock \bibinfo{journal}{\emph{Information storage and retrieval}} \bibinfo{volume}{9}, \bibinfo{number}{11} (\bibinfo{year}{1973}), \bibinfo{pages}{619--633}.
\newblock


\bibitem[Joshi et~al\mbox{.}(2017)]%
        {joshi2017triviaqa}
\bibfield{author}{\bibinfo{person}{Mandar Joshi}, \bibinfo{person}{Eunsol Choi}, \bibinfo{person}{Daniel Weld}, {and} \bibinfo{person}{Luke Zettlemoyer}.} \bibinfo{year}{2017}\natexlab{}.
\newblock \showarticletitle{TriviaQA: A Large Scale Distantly Supervised Challenge Dataset for Reading Comprehension}. In \bibinfo{booktitle}{\emph{Proceedings of the 55th Annual Meeting of the Association for Computational Linguistics (Volume 1: Long Papers)}}. Association for Computational Linguistics.
\newblock


\bibitem[Koukounas et~al\mbox{.}(2024)]%
        {koukounas2024jina}
\bibfield{author}{\bibinfo{person}{Andreas Koukounas}, \bibinfo{person}{Georgios Mastrapas}, \bibinfo{person}{Bo Wang}, \bibinfo{person}{Mohammad~Kalim Akram}, \bibinfo{person}{Sedigheh Eslami}, \bibinfo{person}{Michael G{\"u}nther}, \bibinfo{person}{Isabelle Mohr}, \bibinfo{person}{Saba Sturua}, \bibinfo{person}{Scott Martens}, \bibinfo{person}{Nan Wang}, {et~al\mbox{.}}} \bibinfo{year}{2024}\natexlab{}.
\newblock \showarticletitle{jina-clip-v2: Multilingual Multimodal Embeddings for Text and Images}.
\newblock \bibinfo{journal}{\emph{arXiv preprint arXiv:2412.08802}} (\bibinfo{year}{2024}).
\newblock


\bibitem[Kwiatkowski et~al\mbox{.}(2019)]%
        {Naturalquestions}
\bibfield{author}{\bibinfo{person}{Tom Kwiatkowski}, \bibinfo{person}{Jennimaria Palomaki}, \bibinfo{person}{Olivia Redfield}, \bibinfo{person}{Michael Collins}, \bibinfo{person}{Ankur Parikh}, \bibinfo{person}{Chris Alberti}, \bibinfo{person}{Danielle Epstein}, \bibinfo{person}{Illia Polosukhin}, \bibinfo{person}{Jacob Devlin}, \bibinfo{person}{Kenton Lee}, {et~al\mbox{.}}} \bibinfo{year}{2019}\natexlab{}.
\newblock \showarticletitle{Natural questions: a benchmark for question answering research}.
\newblock \bibinfo{journal}{\emph{Transactions of the Association for Computational Linguistics}}  \bibinfo{volume}{7} (\bibinfo{year}{2019}), \bibinfo{pages}{453--466}.
\newblock


\bibitem[Lewis et~al\mbox{.}(2020)]%
        {2020RAG}
\bibfield{author}{\bibinfo{person}{Patrick S.~H. Lewis}, \bibinfo{person}{Ethan Perez}, \bibinfo{person}{Aleksandra Piktus}, {et~al\mbox{.}}} \bibinfo{year}{2020}\natexlab{}.
\newblock \showarticletitle{Retrieval-Augmented Generation for Knowledge-Intensive {NLP} Tasks}. In \bibinfo{booktitle}{\emph{NeurIPS}}.
\newblock


\bibitem[Lin(2004)]%
        {lin-2004-rouge}
\bibfield{author}{\bibinfo{person}{Chin-Yew Lin}.} \bibinfo{year}{2004}\natexlab{}.
\newblock \showarticletitle{{ROUGE}: A Package for Automatic Evaluation of Summaries}. In \bibinfo{booktitle}{\emph{Text Summarization Branches Out}}. \bibinfo{publisher}{Association for Computational Linguistics}, \bibinfo{address}{Barcelona, Spain}, \bibinfo{pages}{74--81}.
\newblock
\urldef\tempurl%
\url{https://aclanthology.org/W04-1013/}
\showURL{%
\tempurl}


\bibitem[Liu et~al\mbox{.}(2024)]%
        {liu2024deepseek}
\bibfield{author}{\bibinfo{person}{Aixin Liu}, \bibinfo{person}{Bei Feng}, \bibinfo{person}{Bing Xue}, \bibinfo{person}{Bingxuan Wang}, \bibinfo{person}{Bochao Wu}, \bibinfo{person}{Chengda Lu}, \bibinfo{person}{Chenggang Zhao}, \bibinfo{person}{Chengqi Deng}, \bibinfo{person}{Chenyu Zhang}, \bibinfo{person}{Chong Ruan}, {et~al\mbox{.}}} \bibinfo{year}{2024}\natexlab{}.
\newblock \showarticletitle{DeepSeek-V3 Technical Report}.
\newblock \bibinfo{journal}{\emph{arXiv preprint arXiv:2412.19437}} (\bibinfo{year}{2024}).
\newblock


\bibitem[Liu et~al\mbox{.}(2023)]%
        {liu2023learning}
\bibfield{author}{\bibinfo{person}{Haotian Liu}, \bibinfo{person}{Kilho Son}, \bibinfo{person}{Jianwei Yang}, \bibinfo{person}{Ce Liu}, \bibinfo{person}{Jianfeng Gao}, \bibinfo{person}{Yong~Jae Lee}, {and} \bibinfo{person}{Chunyuan Li}.} \bibinfo{year}{2023}\natexlab{}.
\newblock \showarticletitle{Learning customized visual models with retrieval-augmented knowledge}. In \bibinfo{booktitle}{\emph{Proceedings of the IEEE/CVF Conference on Computer Vision and Pattern Recognition}}. \bibinfo{pages}{15148--15158}.
\newblock


\bibitem[Liu(2022)]%
        {Liu_LlamaIndex_2022}
\bibfield{author}{\bibinfo{person}{Jerry Liu}.} \bibinfo{year}{2022}\natexlab{}.
\newblock \bibinfo{booktitle}{\emph{{LlamaIndex}}}.
\newblock
\href{https://doi.org/10.5281/zenodo.1234}{doi:\nolinkurl{10.5281/zenodo.1234}}


\bibitem[Ma et~al\mbox{.}(2024)]%
        {ma2024multi}
\bibfield{author}{\bibinfo{person}{Zi-Ao Ma}, \bibinfo{person}{Tian Lan}, \bibinfo{person}{Rong-Cheng Tu}, \bibinfo{person}{Yong Hu}, \bibinfo{person}{Heyan Huang}, {and} \bibinfo{person}{Xian-Ling Mao}.} \bibinfo{year}{2024}\natexlab{}.
\newblock \showarticletitle{Multi-modal Retrieval Augmented Multi-modal Generation: A Benchmark, Evaluate Metrics and Strong Baselines}.
\newblock \bibinfo{journal}{\emph{arXiv preprint arXiv:2411.16365}} (\bibinfo{year}{2024}).
\newblock


\bibitem[Marino et~al\mbox{.}(2019)]%
        {marino2019ok}
\bibfield{author}{\bibinfo{person}{Kenneth Marino}, \bibinfo{person}{Mohammad Rastegari}, \bibinfo{person}{Ali Farhadi}, {and} \bibinfo{person}{Roozbeh Mottaghi}.} \bibinfo{year}{2019}\natexlab{}.
\newblock \showarticletitle{Ok-vqa: A visual question answering benchmark requiring external knowledge}. In \bibinfo{booktitle}{\emph{Proceedings of the IEEE/cvf conference on computer vision and pattern recognition}}. \bibinfo{pages}{3195--3204}.
\newblock


\bibitem[Nguyen et~al\mbox{.}(2016)]%
        {MSMARCO}
\bibfield{author}{\bibinfo{person}{Tri Nguyen}, \bibinfo{person}{Mir Rosenberg}, \bibinfo{person}{Xia Song}, \bibinfo{person}{Jianfeng Gao}, \bibinfo{person}{Saurabh Tiwary}, \bibinfo{person}{Rangan Majumder}, {and} \bibinfo{person}{Li Deng}.} \bibinfo{year}{2016}\natexlab{}.
\newblock \showarticletitle{MS MARCO: A Human Generated MAchine Reading COmprehension Dataset}.
\newblock \bibinfo{journal}{\emph{choice}}  \bibinfo{volume}{2640} (\bibinfo{year}{2016}), \bibinfo{pages}{660}.
\newblock


\bibitem[OpenAI(2024a)]%
        {gpt4o_mini}
\bibfield{author}{\bibinfo{person}{OpenAI}.} \bibinfo{year}{2024}\natexlab{a}.
\newblock \showarticletitle{GPT-4o mini: advancing cost-efficient intelligence}.
\newblock \bibinfo{journal}{\emph{OpenAI Blog}} (\bibinfo{year}{2024}).
\newblock
\urldef\tempurl%
\url{https://openai.com/index/gpt-4o-mini-advancing-cost-efficient-intelligence//}
\showURL{%
\tempurl}


\bibitem[OpenAI(2024b)]%
        {gpt4o}
\bibfield{author}{\bibinfo{person}{OpenAI}.} \bibinfo{year}{2024}\natexlab{b}.
\newblock \showarticletitle{Hello GPT-4o}.
\newblock \bibinfo{journal}{\emph{OpenAI Blog}} (\bibinfo{year}{2024}).
\newblock
\urldef\tempurl%
\url{https://openai.com/index/hello-gpt-4o/}
\showURL{%
\tempurl}


\bibitem[Papineni et~al\mbox{.}(2002)]%
        {papineni2002bleu}
\bibfield{author}{\bibinfo{person}{Kishore Papineni}, \bibinfo{person}{Salim Roukos}, \bibinfo{person}{Todd Ward}, {and} \bibinfo{person}{Wei-Jing Zhu}.} \bibinfo{year}{2002}\natexlab{}.
\newblock \showarticletitle{Bleu: a method for automatic evaluation of machine translation}. In \bibinfo{booktitle}{\emph{Proceedings of the 40th annual meeting of the Association for Computational Linguistics}}. \bibinfo{pages}{311--318}.
\newblock


\bibitem[Petroni et~al\mbox{.}(2020)]%
        {petroni2020kilt}
\bibfield{author}{\bibinfo{person}{Fabio Petroni}, \bibinfo{person}{Aleksandra Piktus}, \bibinfo{person}{Angela Fan}, \bibinfo{person}{Patrick Lewis}, \bibinfo{person}{Majid Yazdani}, \bibinfo{person}{Nicola De~Cao}, \bibinfo{person}{James Thorne}, \bibinfo{person}{Yacine Jernite}, \bibinfo{person}{Vladimir Karpukhin}, \bibinfo{person}{Jean Maillard}, {et~al\mbox{.}}} \bibinfo{year}{2020}\natexlab{}.
\newblock \showarticletitle{KILT: a benchmark for knowledge intensive language tasks}.
\newblock \bibinfo{journal}{\emph{arXiv preprint arXiv:2009.02252}} (\bibinfo{year}{2020}).
\newblock


\bibitem[Radford et~al\mbox{.}(2021)]%
        {radford2021learning}
\bibfield{author}{\bibinfo{person}{Alec Radford}, \bibinfo{person}{Jong~Wook Kim}, \bibinfo{person}{Chris Hallacy}, \bibinfo{person}{Aditya Ramesh}, \bibinfo{person}{Gabriel Goh}, \bibinfo{person}{Sandhini Agarwal}, \bibinfo{person}{Girish Sastry}, \bibinfo{person}{Amanda Askell}, \bibinfo{person}{Pamela Mishkin}, \bibinfo{person}{Jack Clark}, {et~al\mbox{.}}} \bibinfo{year}{2021}\natexlab{}.
\newblock \showarticletitle{Learning transferable visual models from natural language supervision}. In \bibinfo{booktitle}{\emph{International conference on machine learning}}. PMLR, \bibinfo{pages}{8748--8763}.
\newblock


\bibitem[Rajpurkar et~al\mbox{.}(2016)]%
        {rajpurkar2016squad}
\bibfield{author}{\bibinfo{person}{Pranav Rajpurkar}, \bibinfo{person}{Jian Zhang}, \bibinfo{person}{Konstantin Lopyrev}, {and} \bibinfo{person}{Percy Liang}.} \bibinfo{year}{2016}\natexlab{}.
\newblock \showarticletitle{SQuAD: 100,000+ Questions for Machine Comprehension of Text}. In \bibinfo{booktitle}{\emph{Proceedings of the 2016 Conference on Empirical Methods in Natural Language Processing}}. Association for Computational Linguistics.
\newblock


\bibitem[Schwenk et~al\mbox{.}(2022)]%
        {schwenk2022okvqa}
\bibfield{author}{\bibinfo{person}{Dustin Schwenk}, \bibinfo{person}{Apoorv Khandelwal}, \bibinfo{person}{Christopher Clark}, \bibinfo{person}{Kenneth Marino}, {and} \bibinfo{person}{Roozbeh Mottaghi}.} \bibinfo{year}{2022}\natexlab{}.
\newblock \showarticletitle{A-okvqa: A benchmark for visual question answering using world knowledge}. In \bibinfo{booktitle}{\emph{European conference on computer vision}}. Springer, \bibinfo{pages}{146--162}.
\newblock


\bibitem[Shrivastava and Li(2014)]%
        {shrivastava2014defense}
\bibfield{author}{\bibinfo{person}{Anshumali Shrivastava} {and} \bibinfo{person}{Ping Li}.} \bibinfo{year}{2014}\natexlab{}.
\newblock \showarticletitle{In defense of minhash over simhash}. In \bibinfo{booktitle}{\emph{Artificial Intelligence and Statistics}}. PMLR, \bibinfo{pages}{886--894}.
\newblock


\bibitem[Srinivasan et~al\mbox{.}(2021)]%
        {srinivasan2021wit}
\bibfield{author}{\bibinfo{person}{Krishna Srinivasan}, \bibinfo{person}{Karthik Raman}, \bibinfo{person}{Jiecao Chen}, \bibinfo{person}{Michael Bendersky}, {and} \bibinfo{person}{Marc Najork}.} \bibinfo{year}{2021}\natexlab{}.
\newblock \showarticletitle{Wit: Wikipedia-based image text dataset for multimodal multilingual machine learning}. In \bibinfo{booktitle}{\emph{Proceedings of the 44th international ACM SIGIR conference on research and development in information retrieval}}. \bibinfo{pages}{2443--2449}.
\newblock


\bibitem[Talmor et~al\mbox{.}(2021)]%
        {talmor2021multimodalqa}
\bibfield{author}{\bibinfo{person}{Alon Talmor}, \bibinfo{person}{Ori Yoran}, \bibinfo{person}{Amnon Catav}, \bibinfo{person}{Dan Lahav}, \bibinfo{person}{Yizhong Wang}, \bibinfo{person}{Akari Asai}, \bibinfo{person}{Gabriel Ilharco}, \bibinfo{person}{Hannaneh Hajishirzi}, {and} \bibinfo{person}{Jonathan Berant}.} \bibinfo{year}{2021}\natexlab{}.
\newblock \showarticletitle{Multimodalqa: Complex question answering over text, tables and images}.
\newblock \bibinfo{journal}{\emph{arXiv preprint arXiv:2104.06039}} (\bibinfo{year}{2021}).
\newblock


\bibitem[Team et~al\mbox{.}(2024)]%
        {team2024gemini}
\bibfield{author}{\bibinfo{person}{Gemini Team}, \bibinfo{person}{Petko Georgiev}, \bibinfo{person}{Ving~Ian Lei}, \bibinfo{person}{Ryan Burnell}, \bibinfo{person}{Libin Bai}, \bibinfo{person}{Anmol Gulati}, \bibinfo{person}{Garrett Tanzer}, \bibinfo{person}{Damien Vincent}, \bibinfo{person}{Zhufeng Pan}, \bibinfo{person}{Shibo Wang}, {et~al\mbox{.}}} \bibinfo{year}{2024}\natexlab{}.
\newblock \showarticletitle{Gemini 1.5: Unlocking multimodal understanding across millions of tokens of context}.
\newblock \bibinfo{journal}{\emph{arXiv preprint arXiv:2403.05530}} (\bibinfo{year}{2024}).
\newblock


\bibitem[Wang et~al\mbox{.}(2024c)]%
        {wang2024mineru}
\bibfield{author}{\bibinfo{person}{Bin Wang}, \bibinfo{person}{Chao Xu}, \bibinfo{person}{Xiaomeng Zhao}, \bibinfo{person}{Linke Ouyang}, \bibinfo{person}{Fan Wu}, \bibinfo{person}{Zhiyuan Zhao}, \bibinfo{person}{Rui Xu}, \bibinfo{person}{Kaiwen Liu}, \bibinfo{person}{Yuan Qu}, \bibinfo{person}{Fukai Shang}, {et~al\mbox{.}}} \bibinfo{year}{2024}\natexlab{c}.
\newblock \showarticletitle{Mineru: An open-source solution for precise document content extraction}.
\newblock \bibinfo{journal}{\emph{arXiv preprint arXiv:2409.18839}} (\bibinfo{year}{2024}).
\newblock


\bibitem[Wang et~al\mbox{.}(2024b)]%
        {wang2024comprehensive}
\bibfield{author}{\bibinfo{person}{Jiaqi Wang}, \bibinfo{person}{Hanqi Jiang}, \bibinfo{person}{Yiheng Liu}, \bibinfo{person}{Chong Ma}, \bibinfo{person}{Xu Zhang}, \bibinfo{person}{Yi Pan}, \bibinfo{person}{Mengyuan Liu}, \bibinfo{person}{Peiran Gu}, \bibinfo{person}{Sichen Xia}, \bibinfo{person}{Wenjun Li}, {et~al\mbox{.}}} \bibinfo{year}{2024}\natexlab{b}.
\newblock \showarticletitle{A comprehensive review of multimodal large language models: Performance and challenges across different tasks}.
\newblock \bibinfo{journal}{\emph{arXiv preprint arXiv:2408.01319}} (\bibinfo{year}{2024}).
\newblock


\bibitem[Wang et~al\mbox{.}(2024a)]%
        {wang2024qwen2}
\bibfield{author}{\bibinfo{person}{Peng Wang}, \bibinfo{person}{Shuai Bai}, \bibinfo{person}{Sinan Tan}, \bibinfo{person}{Shijie Wang}, \bibinfo{person}{Zhihao Fan}, \bibinfo{person}{Jinze Bai}, \bibinfo{person}{Keqin Chen}, \bibinfo{person}{Xuejing Liu}, \bibinfo{person}{Jialin Wang}, \bibinfo{person}{Wenbin Ge}, {et~al\mbox{.}}} \bibinfo{year}{2024}\natexlab{a}.
\newblock \showarticletitle{Qwen2-vl: Enhancing vision-language model's perception of the world at any resolution}.
\newblock \bibinfo{journal}{\emph{arXiv preprint arXiv:2409.12191}} (\bibinfo{year}{2024}).
\newblock


\bibitem[Wang et~al\mbox{.}(2024d)]%
        {QAE}
\bibfield{author}{\bibinfo{person}{Zhengren Wang}, \bibinfo{person}{Qinhan Yu}, \bibinfo{person}{Shida Wei}, \bibinfo{person}{Zhiyu Li}, \bibinfo{person}{Feiyu Xiong}, \bibinfo{person}{Xiaoxing Wang}, \bibinfo{person}{Simin Niu}, \bibinfo{person}{Hao Liang}, {and} \bibinfo{person}{Wentao Zhang}.} \bibinfo{year}{2024}\natexlab{d}.
\newblock \showarticletitle{QAEncoder: Towards Aligned Representation Learning in Question Answering System}.
\newblock \bibinfo{journal}{\emph{arXiv preprint arXiv:2409.20434}} (\bibinfo{year}{2024}).
\newblock


\bibitem[Wei et~al\mbox{.}(2022)]%
        {COT}
\bibfield{author}{\bibinfo{person}{Jason Wei}, \bibinfo{person}{Xuezhi Wang}, \bibinfo{person}{Dale Schuurmans}, \bibinfo{person}{Maarten Bosma}, \bibinfo{person}{Fei Xia}, \bibinfo{person}{Ed Chi}, \bibinfo{person}{Quoc~V Le}, \bibinfo{person}{Denny Zhou}, {et~al\mbox{.}}} \bibinfo{year}{2022}\natexlab{}.
\newblock \showarticletitle{Chain-of-thought prompting elicits reasoning in large language models}.
\newblock \bibinfo{journal}{\emph{Advances in neural information processing systems}}  \bibinfo{volume}{35} (\bibinfo{year}{2022}), \bibinfo{pages}{24824--24837}.
\newblock


\bibitem[Xiao et~al\mbox{.}(2024)]%
        {BGE}
\bibfield{author}{\bibinfo{person}{Shitao Xiao}, \bibinfo{person}{Zheng Liu}, \bibinfo{person}{Peitian Zhang}, \bibinfo{person}{Niklas Muennighoff}, \bibinfo{person}{Defu Lian}, {and} \bibinfo{person}{Jian-Yun Nie}.} \bibinfo{year}{2024}\natexlab{}.
\newblock \bibinfo{title}{C-Pack: Packaged Resources To Advance General Chinese Embedding}.
\newblock
\showeprint[arxiv]{2309.07597}~[cs.CL]
\urldef\tempurl%
\url{https://arxiv.org/abs/2309.07597}
\showURL{%
\tempurl}


\bibitem[Yagcioglu et~al\mbox{.}(2018)]%
        {yagcioglu2018recipeqa}
\bibfield{author}{\bibinfo{person}{Semih Yagcioglu}, \bibinfo{person}{Aykut Erdem}, \bibinfo{person}{Erkut Erdem}, {and} \bibinfo{person}{Nazli Ikizler-Cinbis}.} \bibinfo{year}{2018}\natexlab{}.
\newblock \showarticletitle{Recipeqa: A challenge dataset for multimodal comprehension of cooking recipes}.
\newblock \bibinfo{journal}{\emph{arXiv preprint arXiv:1809.00812}} (\bibinfo{year}{2018}).
\newblock


\bibitem[Yang et~al\mbox{.}(2018)]%
        {yang2018hotpotqa}
\bibfield{author}{\bibinfo{person}{Zhilin Yang}, \bibinfo{person}{Peng Qi}, \bibinfo{person}{Saizheng Zhang}, \bibinfo{person}{Yoshua Bengio}, \bibinfo{person}{William Cohen}, \bibinfo{person}{Ruslan Salakhutdinov}, {and} \bibinfo{person}{Christopher~D Manning}.} \bibinfo{year}{2018}\natexlab{}.
\newblock \showarticletitle{HotpotQA: A Dataset for Diverse, Explainable Multi-hop Question Answering}. In \bibinfo{booktitle}{\emph{Proceedings of the 2018 Conference on Empirical Methods in Natural Language Processing}}. Association for Computational Linguistics.
\newblock


\bibitem[Zhang et~al\mbox{.}(2023)]%
        {zhang2023internlm}
\bibfield{author}{\bibinfo{person}{Pan Zhang}, \bibinfo{person}{Xiaoyi Dong}, \bibinfo{person}{Bin Wang}, \bibinfo{person}{Yuhang Cao}, \bibinfo{person}{Chao Xu}, \bibinfo{person}{Linke Ouyang}, \bibinfo{person}{Zhiyuan Zhao}, \bibinfo{person}{Haodong Duan}, \bibinfo{person}{Songyang Zhang}, \bibinfo{person}{Shuangrui Ding}, {et~al\mbox{.}}} \bibinfo{year}{2023}\natexlab{}.
\newblock \showarticletitle{Internlm-xcomposer: A vision-language large model for advanced text-image comprehension and composition}.
\newblock \bibinfo{journal}{\emph{arXiv preprint arXiv:2309.15112}} (\bibinfo{year}{2023}).
\newblock


\bibitem[Zhang et~al\mbox{.}(2019)]%
        {zhang2019bertscore}
\bibfield{author}{\bibinfo{person}{Tianyi Zhang}, \bibinfo{person}{Varsha Kishore}, \bibinfo{person}{Felix Wu}, \bibinfo{person}{Kilian~Q Weinberger}, {and} \bibinfo{person}{Yoav Artzi}.} \bibinfo{year}{2019}\natexlab{}.
\newblock \showarticletitle{Bertscore: Evaluating text generation with bert}.
\newblock \bibinfo{journal}{\emph{arXiv preprint arXiv:1904.09675}} (\bibinfo{year}{2019}).
\newblock


\bibitem[Zhao et~al\mbox{.}(2024)]%
        {zhao2024retrieval}
\bibfield{author}{\bibinfo{person}{Penghao Zhao}, \bibinfo{person}{Hailin Zhang}, \bibinfo{person}{Qinhan Yu}, \bibinfo{person}{Zhengren Wang}, \bibinfo{person}{Yunteng Geng}, \bibinfo{person}{Fangcheng Fu}, \bibinfo{person}{Ling Yang}, \bibinfo{person}{Wentao Zhang}, {and} \bibinfo{person}{Bin Cui}.} \bibinfo{year}{2024}\natexlab{}.
\newblock \showarticletitle{Retrieval-augmented generation for ai-generated content: A survey}.
\newblock \bibinfo{journal}{\emph{arXiv preprint arXiv:2402.19473}} (\bibinfo{year}{2024}).
\newblock


\bibitem[Zhu et~al\mbox{.}(2024b)]%
        {zhu2024rageval}
\bibfield{author}{\bibinfo{person}{Kunlun Zhu}, \bibinfo{person}{Yifan Luo}, \bibinfo{person}{Dingling Xu}, \bibinfo{person}{Ruobing Wang}, \bibinfo{person}{Shi Yu}, \bibinfo{person}{Shuo Wang}, \bibinfo{person}{Yukun Yan}, \bibinfo{person}{Zhenghao Liu}, \bibinfo{person}{Xu Han}, \bibinfo{person}{Zhiyuan Liu}, {et~al\mbox{.}}} \bibinfo{year}{2024}\natexlab{b}.
\newblock \showarticletitle{Rageval: Scenario specific rag evaluation dataset generation framework}.
\newblock \bibinfo{journal}{\emph{arXiv preprint arXiv:2408.01262}} (\bibinfo{year}{2024}).
\newblock


\bibitem[Zhu et~al\mbox{.}(2024a)]%
        {zhu2024murar}
\bibfield{author}{\bibinfo{person}{Zhengyuan Zhu}, \bibinfo{person}{Daniel Lee}, \bibinfo{person}{Hong Zhang}, \bibinfo{person}{Sai~Sree Harsha}, \bibinfo{person}{Loic Feujio}, \bibinfo{person}{Akash Maharaj}, {and} \bibinfo{person}{Yunyao Li}.} \bibinfo{year}{2024}\natexlab{a}.
\newblock \showarticletitle{Murar: A simple and effective multimodal retrieval and answer refinement framework for multimodal question answering}.
\newblock \bibinfo{journal}{\emph{arXiv preprint arXiv:2408.08521}} (\bibinfo{year}{2024}).
\newblock


\end{thebibliography}
\end{sloppypar}
\end{document}